
\documentclass{bmvc2k}

\usepackage{amsmath}
\usepackage{amssymb}

\usepackage{pgffor}

\usepackage{graphicx}  

\usepackage{caption}
\usepackage{subcaption}
\usepackage{ulem}
\usepackage{todonotes}
\usepackage[percent]{overpic}

\definecolor{darkorange}{rgb}{0.96, 0.86, 0.19}
\usepackage{soul}

\usepackage[skip=2pt]{caption}

\title{VideoGAN-based Trajectory Proposal for Automated Vehicles}

\addauthor{Annajoyce Mariani}{annajoyce.mariani@tu-berlin.de}{1}
\addauthor{Kira Maag}{kira.maag@hhu.de}{2}
\addauthor{Hanno Gottschalk}{gottschalk@math.tu-berlin.de}{1}

\addinstitution{
 Technical University of Berlin\\
 Berlin, Germany
}
\addinstitution{
Heinrich-Heine-University Düsseldorf \\
Düsseldorf, Germany
}

\runninghead{Mariani et al.}{Videogan-based trajectory proposal}

\begin{document}

\maketitle

\begin{abstract}
Being able to generate realistic trajectory options is at the core of increasing the degree of automation of road vehicles. While model-driven, rule-based, and classical learning-based methods are widely used to tackle these tasks at present, they can struggle to effectively capture the complex, multimodal distributions of future trajectories. In this paper we investigate whether a generative adversarial network (GAN) trained on videos of bird’s-eye view (BEV) traffic scenarios can generate statistically accurate trajectories that correctly capture spatial relationships between the agents. To this end, we propose a pipeline that uses low-resolution BEV occupancy grid videos as training data for a video generative model. From the generated videos of traffic scenarios we extract abstract trajectory data using single-frame object detection and frame-to-frame object matching. We particularly choose a GAN architecture for the fast training and inference times with respect to diffusion models. We obtain our best results within 100 GPU hours of training, with inference times under 20\,ms. We demonstrate the physical realism of the proposed trajectories in terms of distribution alignment of spatial and dynamic parameters with respect to the ground truth videos from the Waymo Open Motion Dataset.
\end{abstract}

\section{Introduction}
\label{sec:intro}

\begin{figure}[t]
    \centering
    \begin{tikzpicture}
    \node[anchor=north west] at (-4.2,0) {\includegraphics[height=8cm]{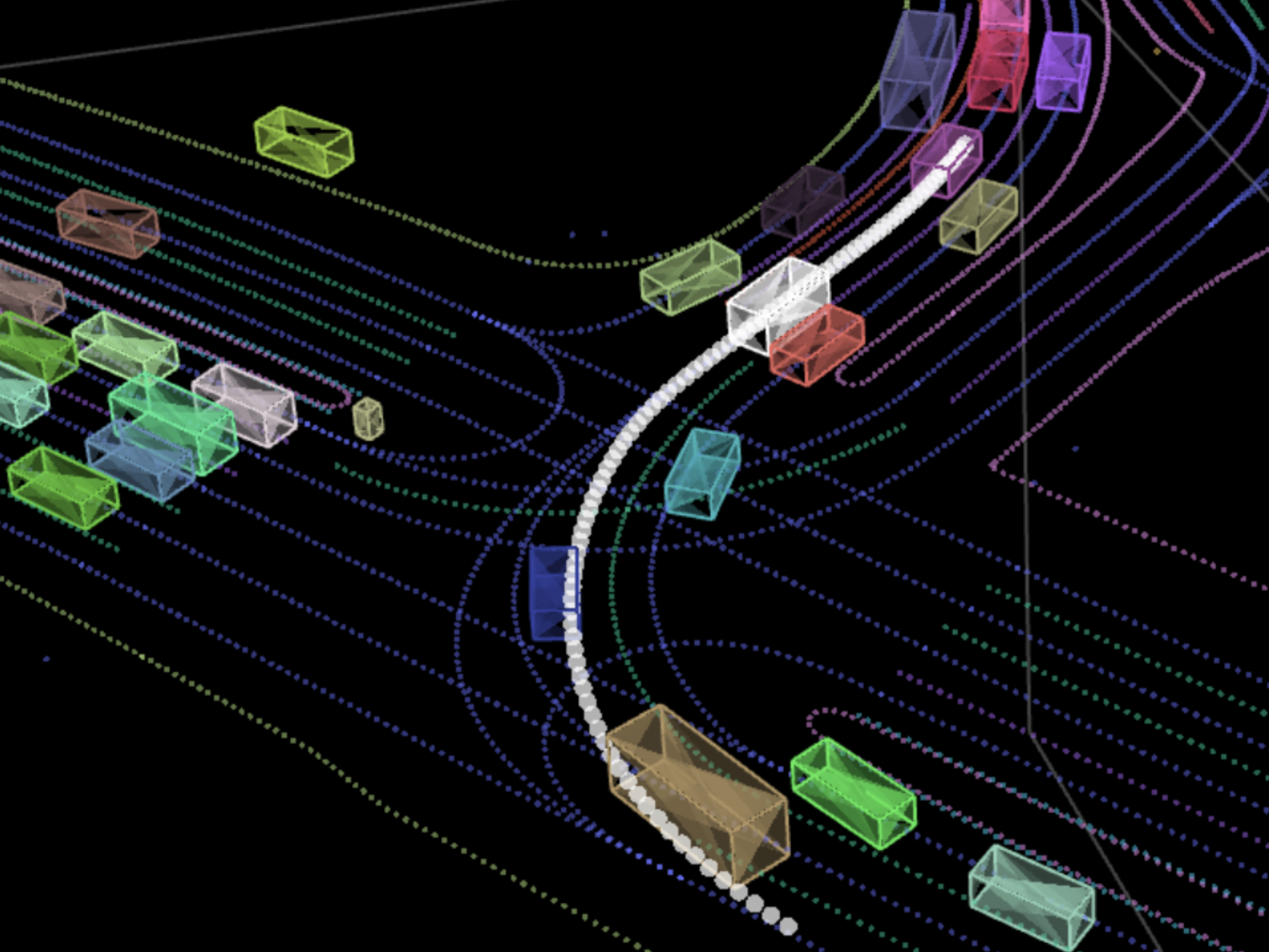}};
    
    \node[anchor=north west] at (7,0) {\includegraphics[height=4.1cm]{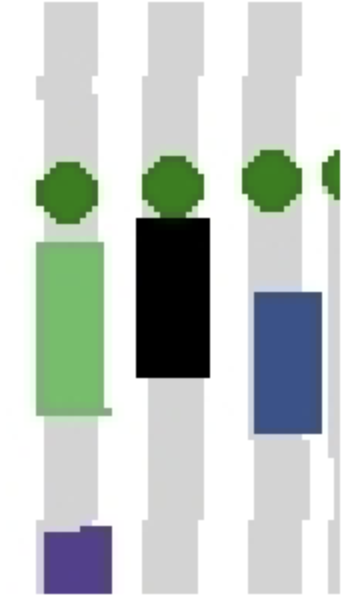}};
    \node[anchor=north west] at (7,-4.4) {\includegraphics[height=3.5cm]{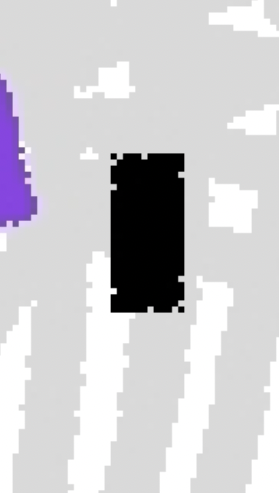}};
    
    \node[anchor=north west] at (9.8,0) {\includegraphics[height=4.1cm]{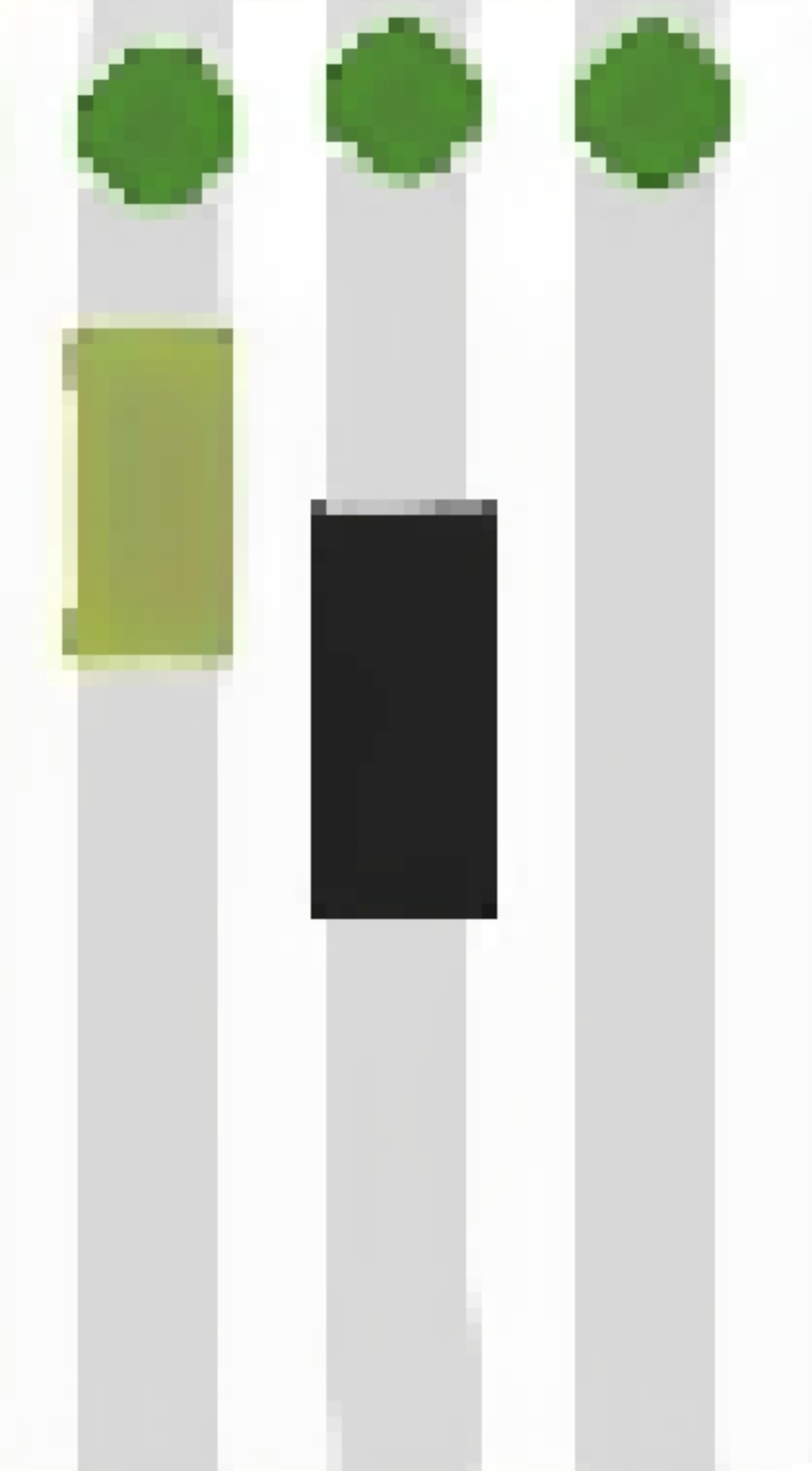}}; 
    \node[anchor=north west] at (9.8,-4.4) {\includegraphics[height=3.5cm]{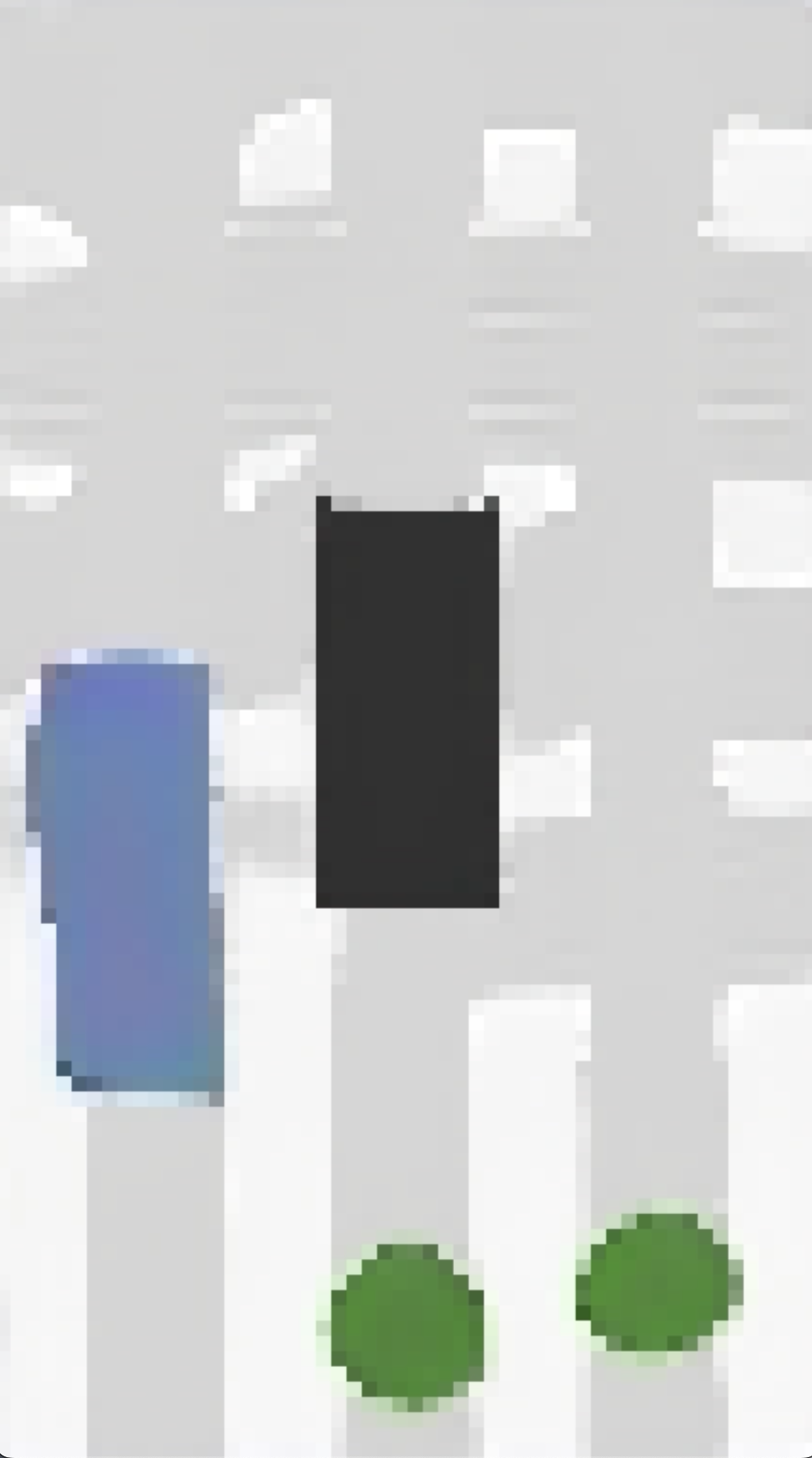}};
\end{tikzpicture}

   \caption{3D representation of a scene from  Waymo Open Motion Dataset (left). Example frames from rasterized BEV occupancy grid videos of real traffic data (center) and generated using our pipeline (right).}
   \label{fig:example}
\end{figure}

Long considered an unattainable goal for its complexity and for the minuscule tolerance to error required, a fully self-driving vehicle would bring countless benefits to our roads: increased safety, reduced traffic congestion, improved fuel efficiency, and higher independence for people with reduced mobility \cite{grigorescu_survey_2020}.
The recent surge in deep learning (DL) capabilities can greatly benefit the field of automated driving (AD) \cite{huang_review_2023, khan_level-5_2023}. The self-driving task is commonly decomposed into the subtasks of perception, prediction, planning, and control \cite{casas_mp3_2021, hu_st-p3_2022}, with trajectory prediction and proposal being pivotal for highly-automated operation of cars, especially in interactive environments such as urban road networks. Trajectory prediction typically involves estimating the future positions of dynamic agents (e.g., vehicles and pedestrians), while planning relies on this information to propose and choose trajectories compliant with the traffic regulations, the safety of other road users and the driving prompts of a human or automated navigator \cite{leon_review_2021, bharilya_machine_2023}.
The modules of this so-called AD stack can either be designed and trained independently or holistically in end-to-end architectures. While all of these tasks can benefit from DL models, prediction and planning of trajectory form the decision-making core of the stack, where the automated agent reasons about future states and selects appropriate actions, based on perceived inputs. Some methods clearly distinguish the two tasks by first predicting possible future paths of road users and later reason over this information to plan a safe route \cite{park_leveraging_2023, cui_lookout_2021}.
Alternative approaches rely on generating realistic futures jointly with a safe motion plan for the ego vehicle, and only intend planning as picking one of these ego trajectories compatibly with the navigation prompts \cite{fang_tpnet_2020, xiong_fine-grained_2025, zhao_novel_2022, hagedorn_integration_2024}.

Classical methods, based on hard-coded logic and rule-based algorithms, can handle simpler driving assistance tasks such as lane keeping, adaptive cruise control or automatic braking \cite{yu_researches_2021, ashwin_deep_2023}, with the main contribution from artificial intelligence being machine vision in the perception module \cite{horgan_vision-based_2015}, but they struggle to model the complex, high-dimensional distributions that govern the joint behavior of multiple agents in high-level automated driving \cite{zare_survey_2023, plaat_deep_2020}.
In contrast, DL models are better candidates to model such complex behavior. On the one hand, prediction is often approached as forecasting a discrete set of individual trajectories, either with attention-based sequential networks \cite{fang_tpnet_2020, ngiam_scene_2021, zhou_hivt_2022}, or more recently with generative models, such as diffusion models or generative adversarial networks (GANs) \cite{jiang_motiondiffuser_2023, li_generative_2025}. These models are better at predicting diverse and multimodal futures, as they are designed to model the full distribution of the data they are trained on by minimizing the distance of the generated distribution from the true data distribution.
On the other hand, approaches modeling scenes from a visual bird's-eye view (BEV) perspective can intrinsically handle spatial layout constraints such as shape-aware safety distances and joint interactions between road users and infrastructure, and contextual cues such as traffic signals and road geometry. These methods also naturally handle trajectories of variable length, such as agents entering and exiting the relevant surrounding of the automated vehicle, which discrete-trajectory methods do not handle gracefully \cite{mahjourian_occupancy_2022, hu_fiery_2021, agro_implicit_2023}.
However, most of such spatially-grounded approaches rely on attention-based architectures, yet do not leverage the aforementioned advantages of generative models 
\cite{hu_fiery_2021, gilles_home_2021, ridel_scene_2020, mahjourian_occupancy_2022}.

In this work, we investigate whether a video generative model trained on low-resolution BEV traffic scenarios can generate statistically accurate trajectories that correctly capture spatial relationships between the agents, leveraging the advantages of both generative and spatially-grounded approaches. Our pipeline for such preliminary trajectory proposal module consists of three part: rasterization of traffic scenes from abstract data, training of the videoGAN model and extraction of trajectories from the generated videos.
Firstly, we extract abstract data relevant to model traffic, i.e. the location of all road users, road center lines and traffic lights, from the Waymo Open Motion Dataset \cite{sun_scalability_2020}, and we rasterize them into a dataset of low-resolution BEV occupancy grid videos.
Secondly, we train a video generative model on the rasterized videos of traffic scenes.
We choose in particular a videoGAN model over a diffusion model due to their significantly lower computational cost and faster training and inference times \cite{xing_survey_2025}.
While diffusion models outperform GANs by design in terms of visual fidelity and sample diversity \cite{xing_survey_2025, saxena_generative_2021}, these advantages are less relevant for the trajectory proposal tasks, where performance is measured by task-specific metrics beyond visual quality, such as physical plausibility or driving feasibility. 
In particular, we choose a videoGAN model that can generate dynamically coherent scenes of any desired length \cite{brooks_generating_2022}. 
Lastly, we extract abstract trajectory data from the generated videos using single-frame object detection and frame-to-frame object matching. 
An example is shown in Figure~\ref{fig:example}. 
To assess the realism of the proposed trajectories, we analyze the distribution alignment of spatial and dynamic parameters such as inter-agent distances and relative speeds with respect to the input data, along with their interactions with traffic lights.
Our code and trained models can be found at https://github.com/ajmariani/video-gan-trajectories.

Our contributions can be summarized as follows: 
\begin{itemize}\setlength\itemsep{-0.5ex}
    \item We present a trajectory proposal method which is both generative and spatially-grounded, with minimal inference times of 20\,ms to generate a 200\,s long video of a traffic scene.
    \item We present a pipeline to rasterize BEV occupancy grid videos from abstract traffic data in a highly customizable way.
    \item We develop a procedure that extracts abstract data on traffic lights and road users from the generated video predictions. 
    \item We demonstrate the ability of video generation models to generate traffic scenes that are realistic and safe in terms of agents interaction and compliance with dynamic traffic signals.
\end{itemize}
\section{Related Work}

\paragraph{Deep learning for trajectory proposal}
Most trajectory proposal approaches formulate the problem as sequentially elaborating a discrete set of possible futures for each agent \cite{chen_trajectory_2025}. 
Examples include encoder-decoder as well as self-attention architectures, such as MTR~\cite{shi_motion_2022}, HiVT~\cite{zhou_hivt_2022}, Scene Transformer\cite{ngiam_scene_2021}, and HDGT~\cite{jia_hdgt_2023}. 
Compared to transformers, generative models are better suited for generating diverse and realistic data from high-dimensional and multimodal distribution as they explicitly learn to approximate the true distribution of the training data \cite{ruthotto_introduction_2021, li_generative_2025}. 
For these reasons, some works have been proposed leveraging generative models, such as diffusion models (e.g., MotionDiffuser~\cite{jiang_motiondiffuser_2023}, MID \cite{gu_stochastic_2022}) or GANs (e.g., SocialGAN~\cite{gupta_social_2018}). 
An alternative approach to trajectory proposal is encoding the scene in a visual format, such as a low-dimensional map or occupancy grid in BEV. This format has the potential to naturally encode spatial layouts and joint contextual cues such as the relative positions and roles of road users as well as static and dynamic map features. Moreover, this bypasses a limitation of the set-of-trajectories format, which typically require sequences of constant length and therefore cannot model agents entering or leaving the surroundings of the decision-making agent.
Such methods are based on LSTMs \cite{ridel_scene_2020}, convolutional networks \cite{hu_fiery_2021, wu_motionnet_2020}, attention blocks and encoder-decoders \cite{mahjourian_occupancy_2022, gilles_home_2021}, and do not leverage the aforementioned benefits of generative models.

Compared to previous works on trajectory proposal, we combine the advantages of generative models and spatially-grounded approaches by presenting a video generative model trained on low-resolution BEV traffic scenarios.

\paragraph{Metrics for trajectory proposal}
Models that predict future trajectories from past trajectories mainly assess prediction accuracy with displacement-based metrics such as the average displacement error, the final displacement error or the miss rate~\cite{jia_hdgt_2023,ngiam_scene_2021,zhou_hivt_2022, shi_motion_2022, alahi_social_2016, salzmann_trajectron_2020}.
However, these metrics are not sufficient to evaluate safety, realism and feasibility, especially when training is performed on synthetic datasets with low physical realism (e.g., CARLA) or limited domain coverage in real road applications (e.g., highways only). Safety related metrics include red-light violations \cite{rhinehart_deep_2019}, collision rate, overlap rate, unsafe spacing, and time spent driving off-road or in the wrong lane \cite{huang_multimodal_2023, rhinehart_deep_2019}. Realism and driving smoothness can be evaluated statistically by comparing the distributions of parameters such as vehicle speed, acceleration, jerk and inter-vehicle distance \cite{jiao_kinematics-aware_2024, karnchanachari_towards_2024}.

We use a similar statistical comparison between the speeds and accelerations of the generated and training data. Since our model proposes scenes of traffic from scratch, we also statistically evaluate traffic density as well as agent shapes and positions. Additionally, we compare speed distributions in the vicinity of green and red traffic lights to assess the model’s compliance with traffic signals.
\begin{figure}[th]
   \centering
   \includegraphics[width=0.98\textwidth]{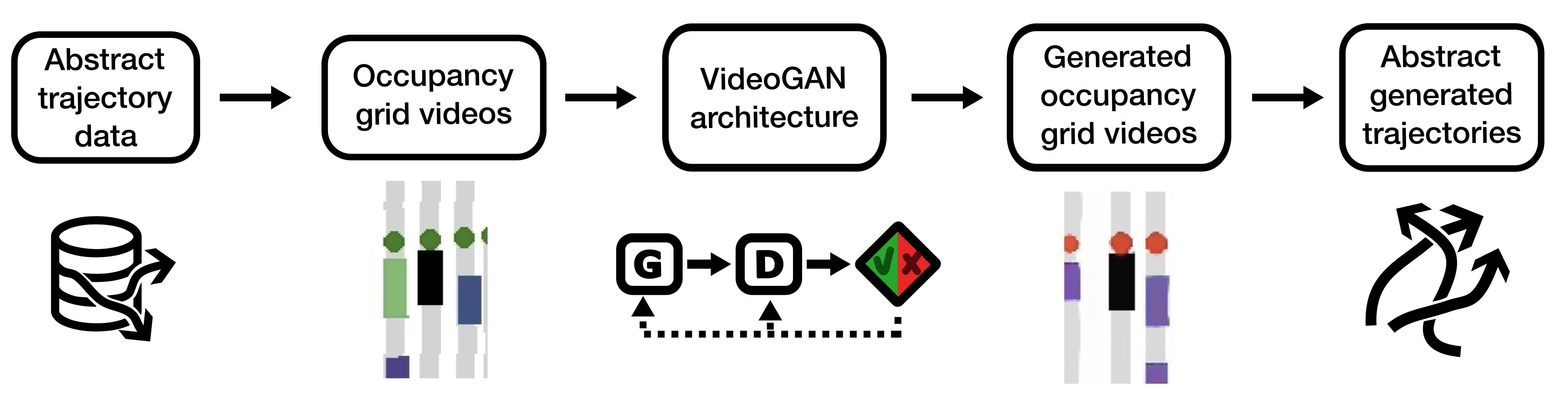}
   \caption{A schematic illustration of our pipeline. We rasterize abstract trajectory data into low-resolution occupancy grid videos of BEV traffic scenes. These videos serve as input to train a videoGAN model, which consists of a generator (G) that attempts to generate realistic data, and a discriminator (D) that distinguishes real data from generated data. Last, we process the generated videos to extract trajectory proposals.}
   \label{fig:method}
\end{figure}

\section{Proposed method}
In this section, we present our pipeline, which consists of rasterizing abstract trajectory data into low-resolution videos of BEV traffic scenes, using these videos as training data for a videoGAN model, and processing the generated videos to extract abstract trajectory data. A schematic illustration is shown in Figure~\ref{fig:method}.

\paragraph{Rasterization of trajectory data}
Our rasterization pipeline requires three types of traffic information: road centerlines, the time-varying positions and states of traffic lights, and the shape and trajectory information for all the agents within a certain surrounding of the ego vehicle, namely the coordinates of the center, the heading and size of the bounding box. 
We convert all the world coordinates into image coordinates, where each scene is centered around and oriented as the ego vehicle from a BEV perspective. Center lanes are rendered as continuous line of specified thickness (here $1.5~\text{m}$) and light gray color in the background. Traffic lights are rendered as a red, green or yellow circle of size $2\times2~\text{m}$ in the foreground. Each road user is rendered as a rectangle of the specified size, position and orientation. The ego car is black, while for other road users the color is randomly sampled from an HSV subspace to avoid overlap with the color of the lanes, the ego car, the white background and the three hues of the traffic lights. Finally, each frame is cropped to window of specified pixel size and aspect ratio that represents the surroundings of the ego vehicle. Each scene is rasterized into a separate occupancy grid video. Example images are shown in Figure~\ref{fig:example} (center).

\begin{figure}[th]
    \centering
    \includegraphics[width=0.14\textwidth]{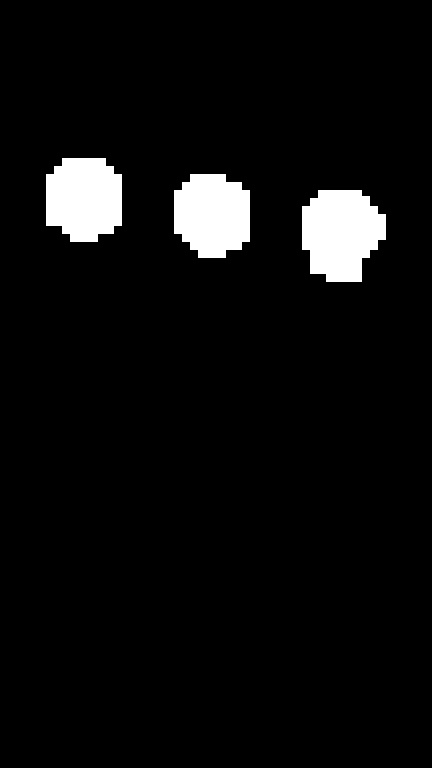}
    \hspace{0.18em}
    \includegraphics[width=0.14\textwidth]{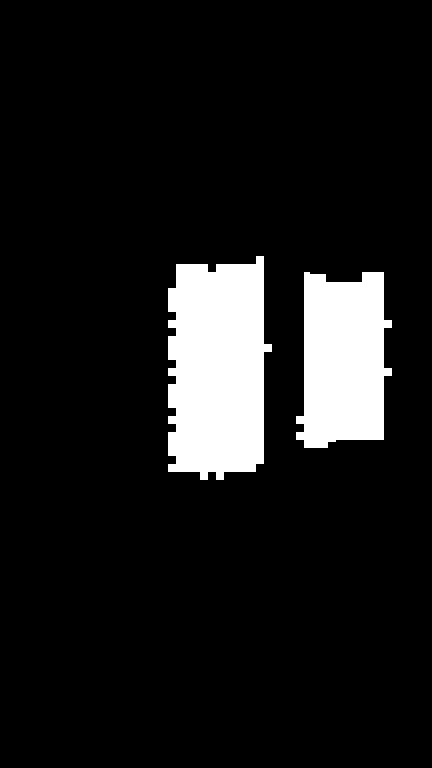}
    \hspace{0.18em}
    \includegraphics[width=0.14\textwidth]{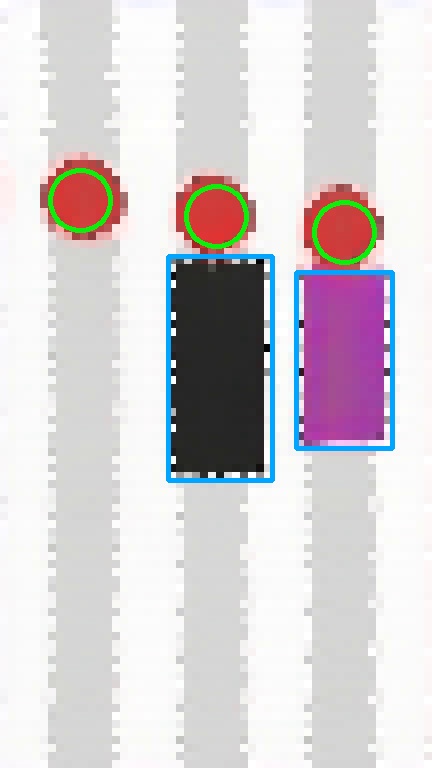}
    \hspace{3ex}
    \includegraphics[width=0.14\textwidth]{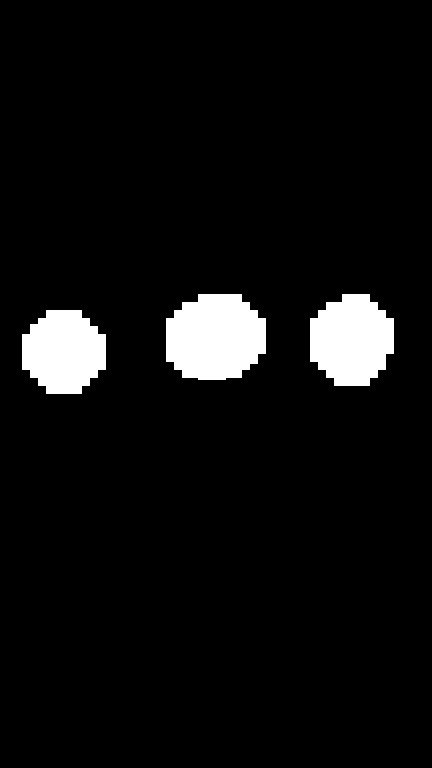}
    \hspace{0.18em}
    \includegraphics[width=0.14\textwidth]{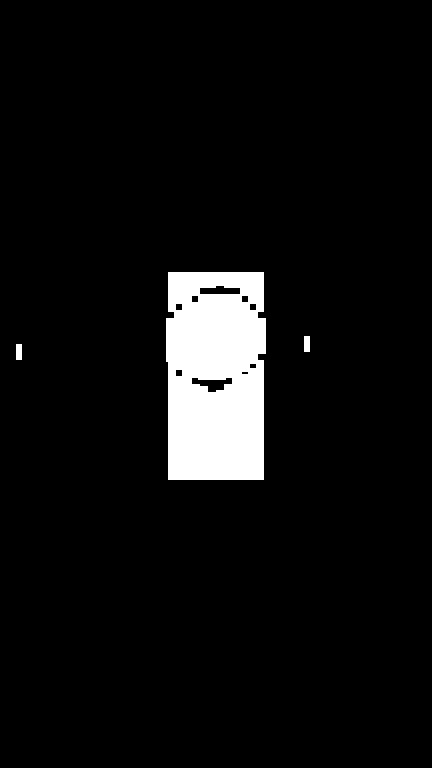}
    \hspace{0.18em}
    \includegraphics[width=0.14\textwidth]{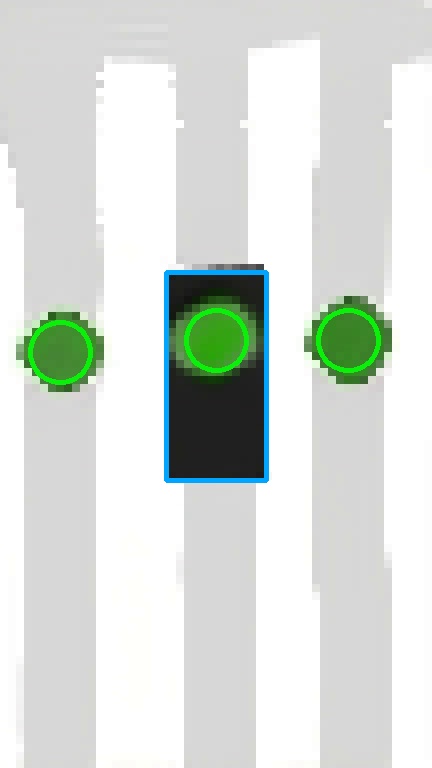}

   \vspace{4pt}
   \caption{Two examples of frames masking and object detection. From left to right, both sequences show first the mask for traffic lights, second the mask for road users, and third the original image overlaid with the pipeline prediction of identity, position and size of the object. Notice that this object detection pipeline is fairly accurate even when a vehicle is near in color or passing beneath a traffic light. In the pipeline the masks for the ego agent and other road users are separated, but they are merged in this figure for visualization purposes.}
   \label{frame_by_frame_decoding}
\end{figure}

\paragraph{Training}
We feed these occupancy grid videos into a generative model, in particular a GAN, to create trajectory proposals. A GAN consists of two neural networks, a generator that attempts to produce realistic data and a discriminator that distinguishes between real and generated data, which are trained against each other until the generator delivers results that are deceptively real. Other generative models could also be incorporated into our pipeline.
We use a videoGAN model which implements spatiotemporal convolutions and latent vectors to generate dynamically coherent scenes of any length \cite{brooks_generating_2022}.
\paragraph{Frame-by-frame object detection from the generated videos}
We first extract information from each frame independently with a frame-by-frame object detection pipeline (see Figure~\ref{frame_by_frame_decoding}). Each frame is converted to HSV color space for improved filtering, and masked in the subspaces around red, green, and yellow, where traffic lights are expected. 
Each traffic light in the image is replaced by the color it would otherwise occlude (e.g., vehicle color or road surface). This procedure improves the detection of agents that are passing beneath a traffic light (see right-hand side of Figure~\ref{frame_by_frame_decoding}).  A similar masking process is applied separately for agents and the ego vehicle.
A step of morphological opening is applied for all three masks to smooth out possible errors at the contours. Each contour is analyzed to extract spatial features, including center coordinates, size, color, area, and shape descriptors such as rectangularity and circularity. For each scene, this pipeline generates a list of objects with these  attributes, along with the corresponding timestep and the mask category from which they were derived. Additionally, an object is reclassified as unknown if its area is inconsistent with the expected size range for its recognized category.

\paragraph{Trajectory extraction}
We match object identities by computing a cost matrix between objects in consecutive frames based on spatial proximity, color similarity, aspect ratio, and type consistency. Persistent IDs are assigned to matched objects, while those lacking a sufficiently accurate match are identified as new agents and assigned a novel ID. Lastly, each object is enriched with instantaneous speed and acceleration relative to the ego agent, computed via interpolation over its trajectory. We assume the same frame rate and spatial scale in the generated videos as in the input data.
\section{Experiments}
\begin{figure}[t]
    \centering

    \begin{minipage}[t]{0.14\textwidth}
        \centering
        \includegraphics[width=\linewidth]{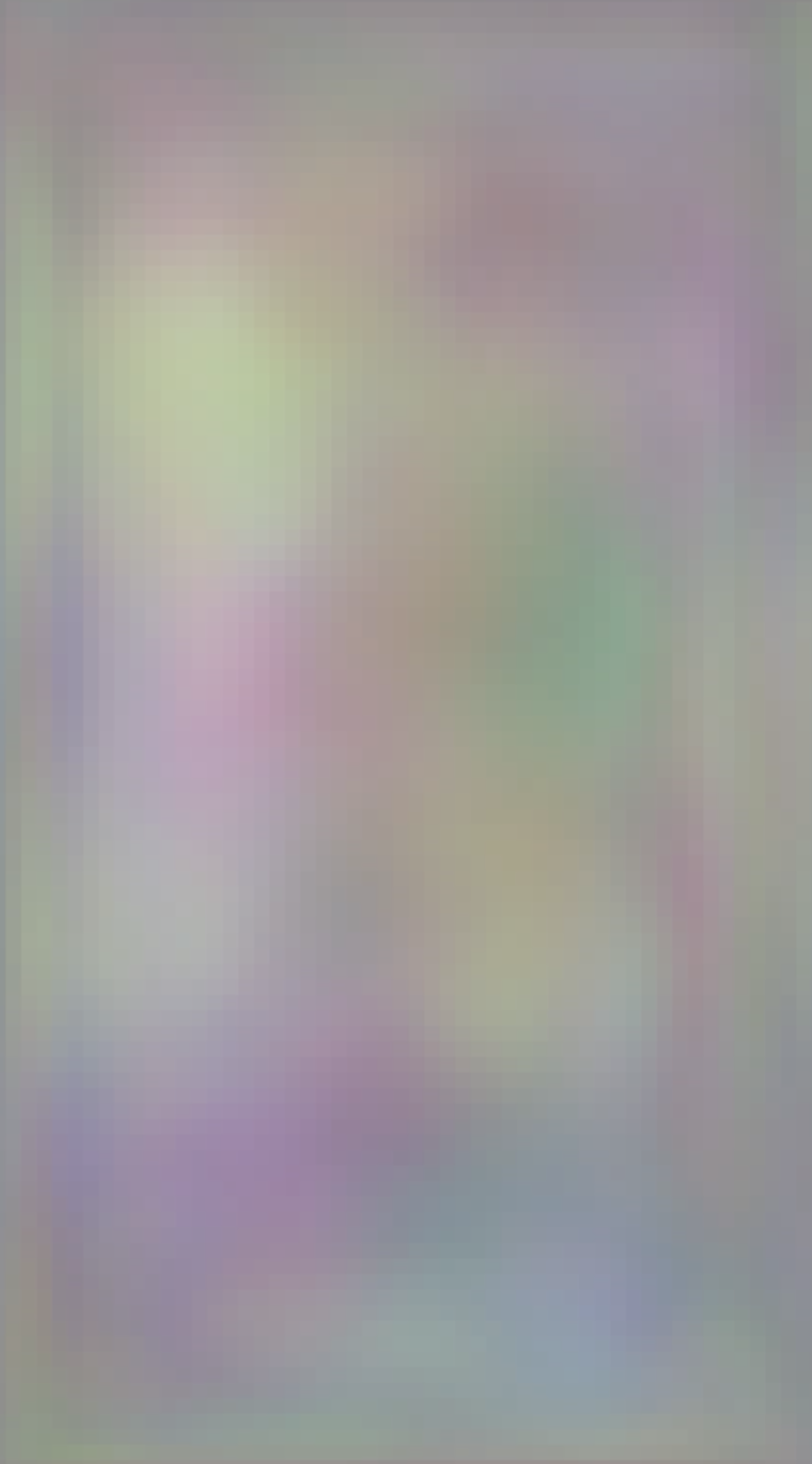}
        \vspace{1ex}
        {\small 0}
    \end{minipage}
    \hspace{0.12em}
    \begin{minipage}[t]{0.14\textwidth}
        \centering
        \includegraphics[width=\linewidth]{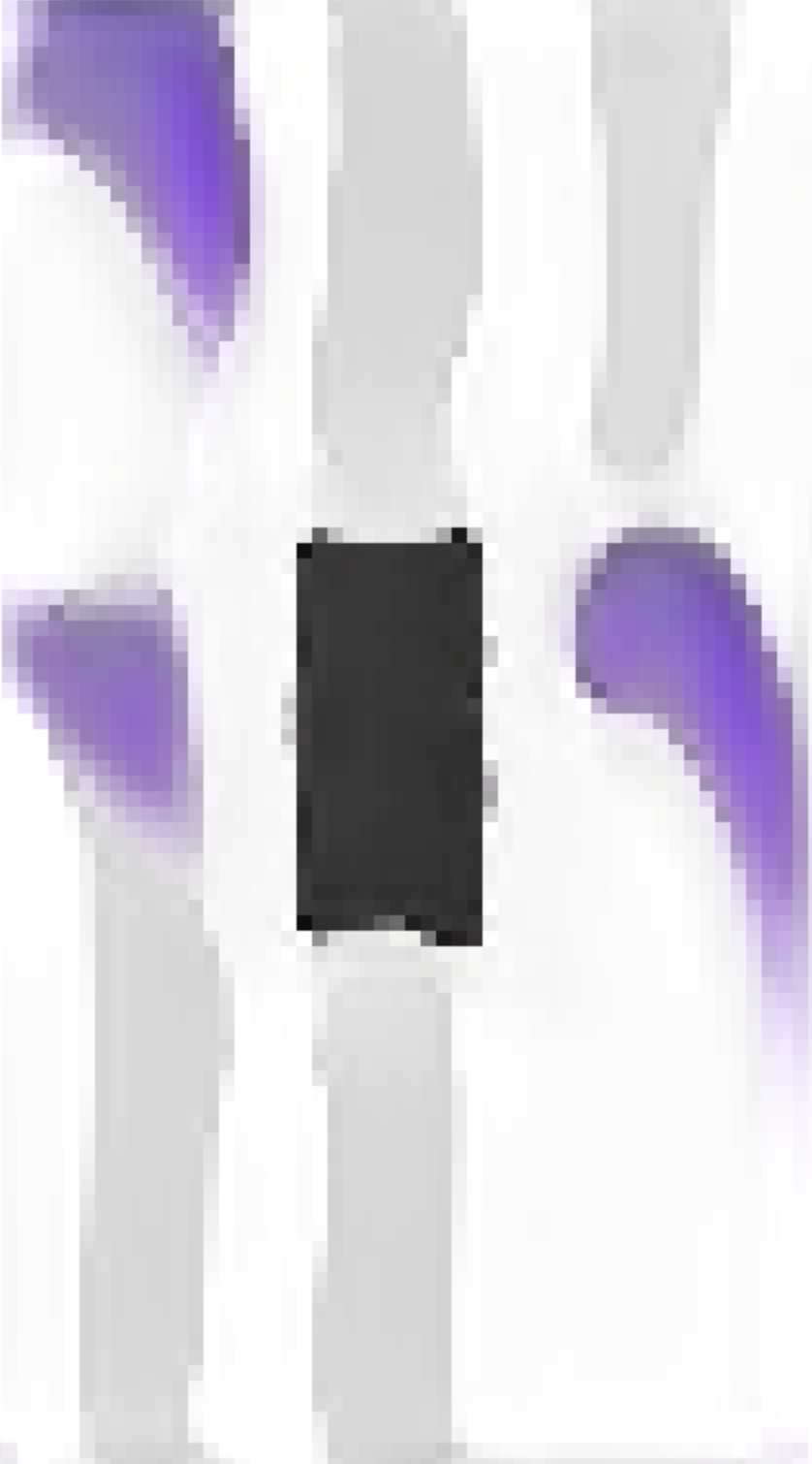}
        \vspace{1ex}
        {\small 10k}
    \end{minipage}
    \hspace{0.12em}
    \begin{minipage}[t]{0.14\textwidth}
        \centering
        \includegraphics[width=\linewidth]{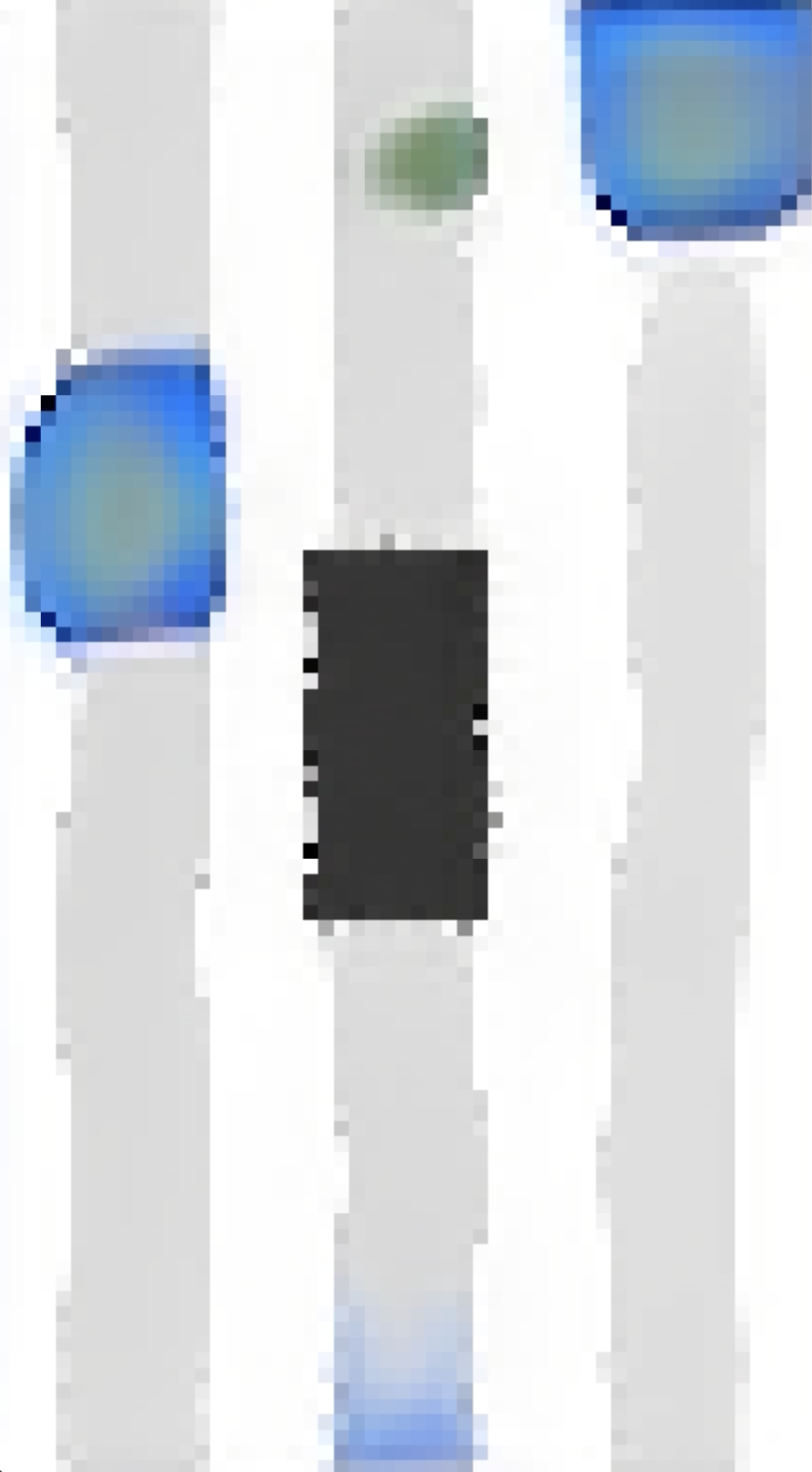}
        \vspace{1ex}
        {\small 20k}
    \end{minipage}
    \hspace{0.12em}
    \begin{minipage}[t]{0.14\textwidth}
        \centering
        \includegraphics[width=\linewidth]{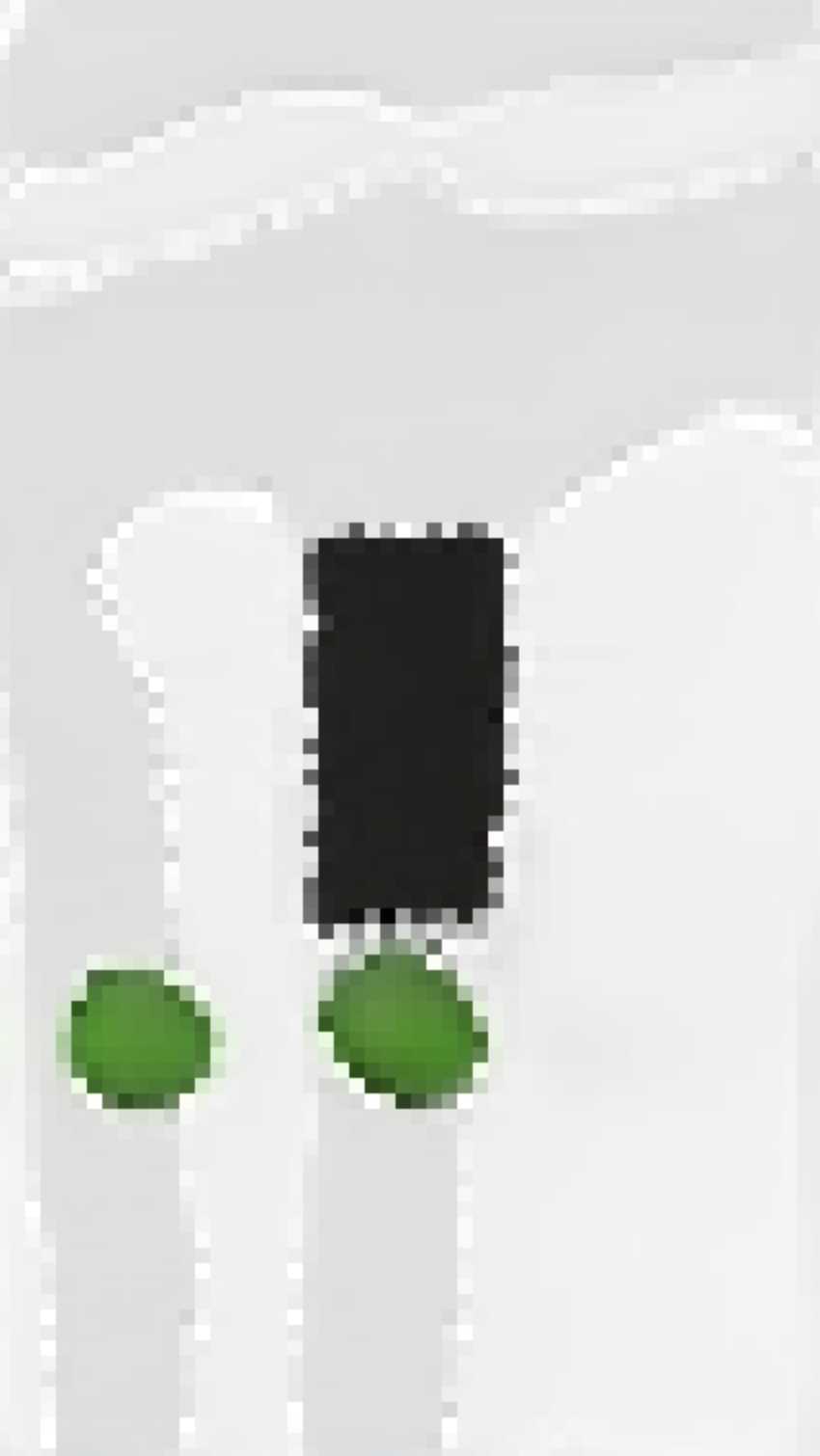}
        \vspace{1ex}
        {\small 40k}
    \end{minipage}
    \hspace{0.12em}
    \begin{minipage}[t]{0.14\textwidth}
        \centering
        \includegraphics[width=\linewidth]{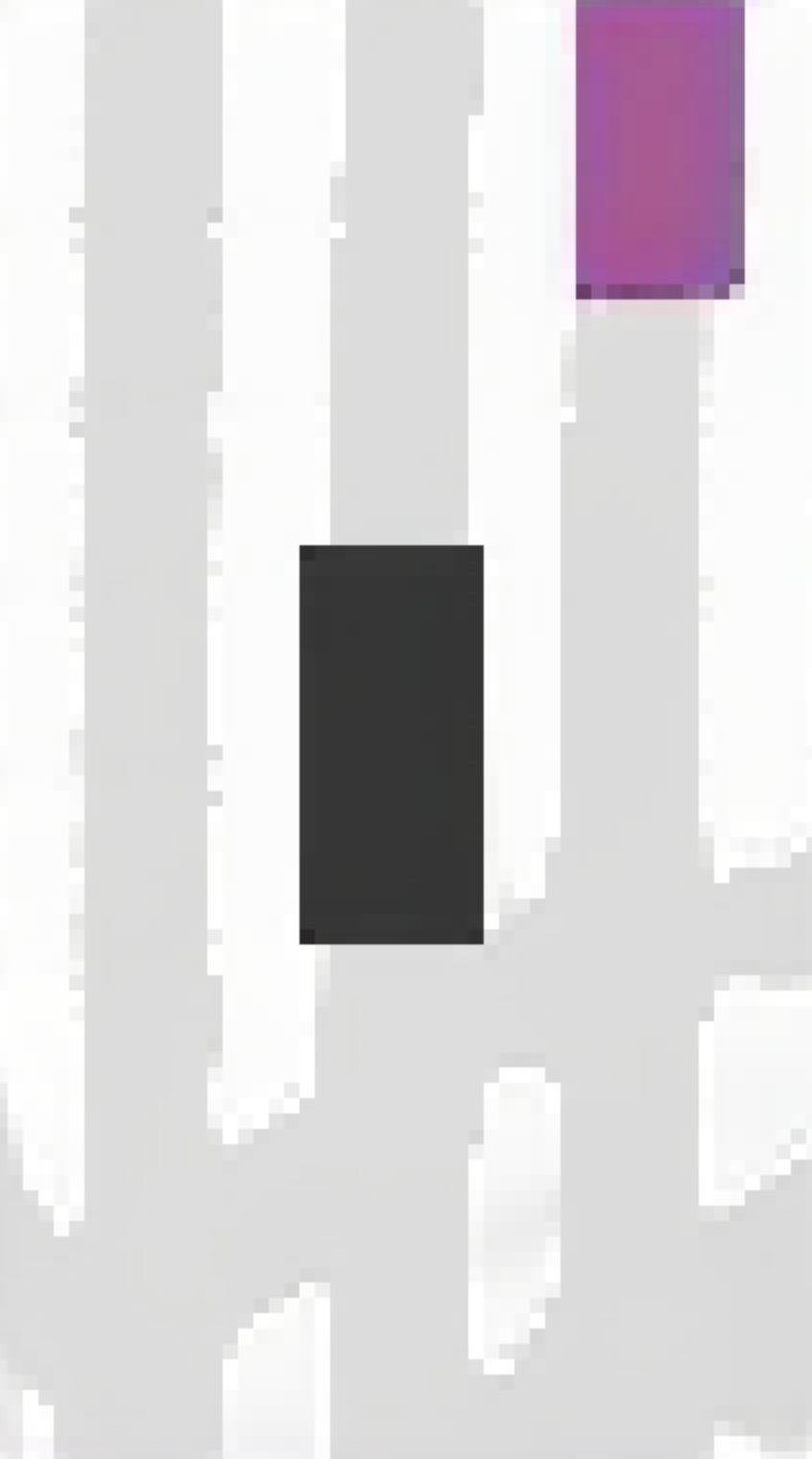}
        \vspace{1ex}
        {\small 70k}
    \end{minipage}
    \hspace{0.12em}
    \begin{minipage}[t]{0.14\textwidth}
        \centering
        \includegraphics[width=\linewidth]{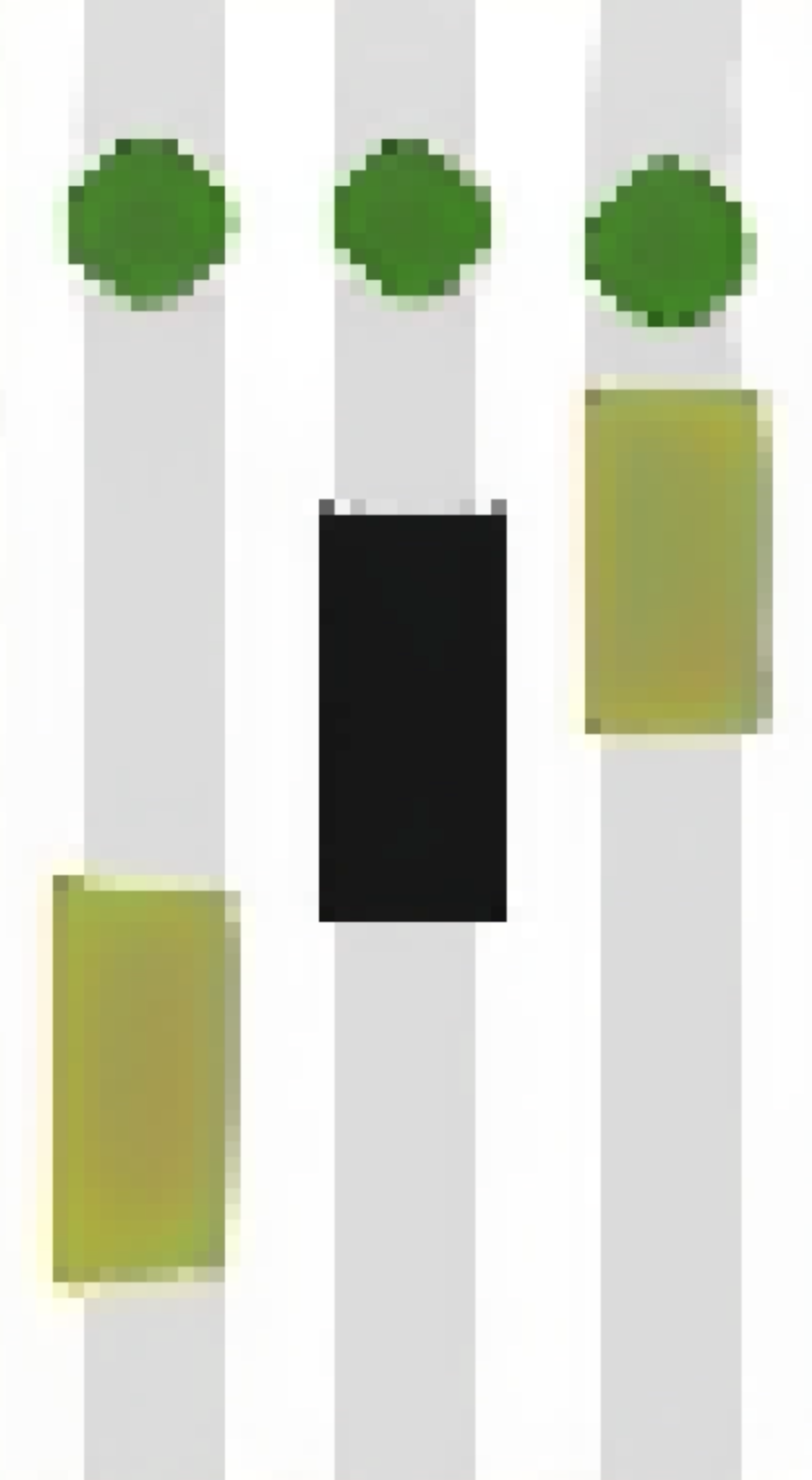}
        \vspace{1ex}
        {\small 90k}
    \end{minipage}
    \caption{Visual example of the training progress for different iterations.}
    \label{fig:train}
\end{figure}

\paragraph{Experimental setting}
We train the low-resolution network from LongVideoGAN \cite{brooks_generating_2022} across 4 40\,GB A100 GPUs for 100\,GPU hours on 20k videos of 15 seconds each, for a total of around 80\,h of driving content. Inference takes approximately 20\,ms for a 15\,s scene and scales linearly with sequence length, reaching 150\,ms for a 2-minute scene. We train for 100k steps with a batch size of 8, as no significant improvement was observed when training on more data, and we prioritize training speed. We monitor the training progress with the Frechet Video Distance, a metric for evaluating the quality of generated videos~\cite{brooks_generating_2022}. The training videos are rasterized from scenes sampled from the Waymo Open Motion Dataset~\cite{sun_scalability_2020}, one of the largest and most diverse driving datasets. It takes around 20\,GPU hours to create the training dataset. When generating the data, we ensured a higher representation of scenes where the ego vehicle turns (around 40\% of the total composition rather than the 10\% naturally featured in Waymo), in order to reflect this behavior in the generated videos. 
For the results presented in the following the window size is chosen of $20\times10$\,m and frame size of $54\times96$ pixels, which corresponds to a scale of around 5\,pixels per meter. Example images of training progress over iterations are shown in Figure~\ref{fig:train}.

\begin{figure}[tb]
    \centering

    \begin{overpic}[width=0.14\textwidth]{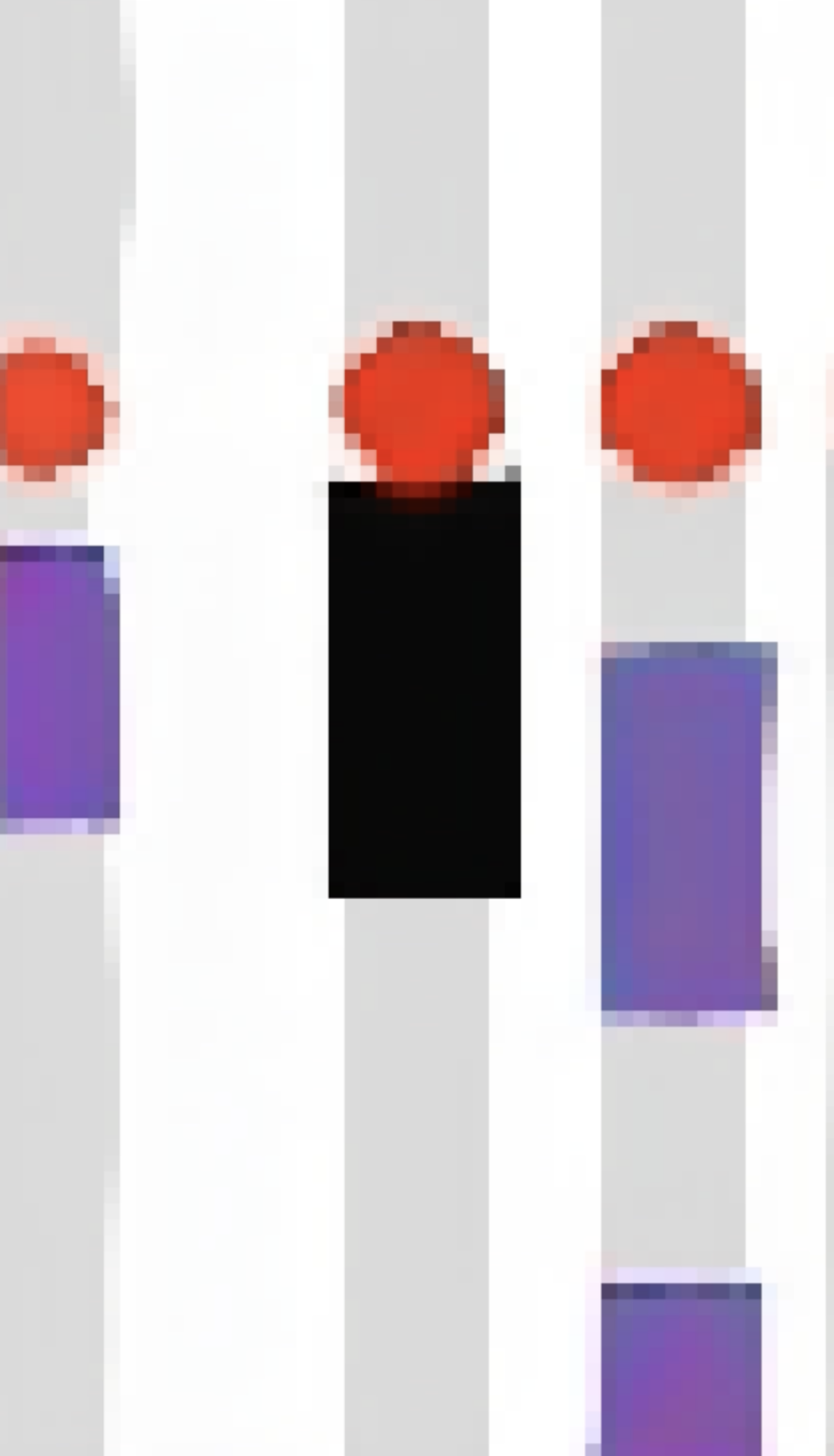}
        \put(21,-12){\small (a)}
    \end{overpic}
    \hspace{0.12em}
    \begin{overpic}[width=0.14\textwidth]{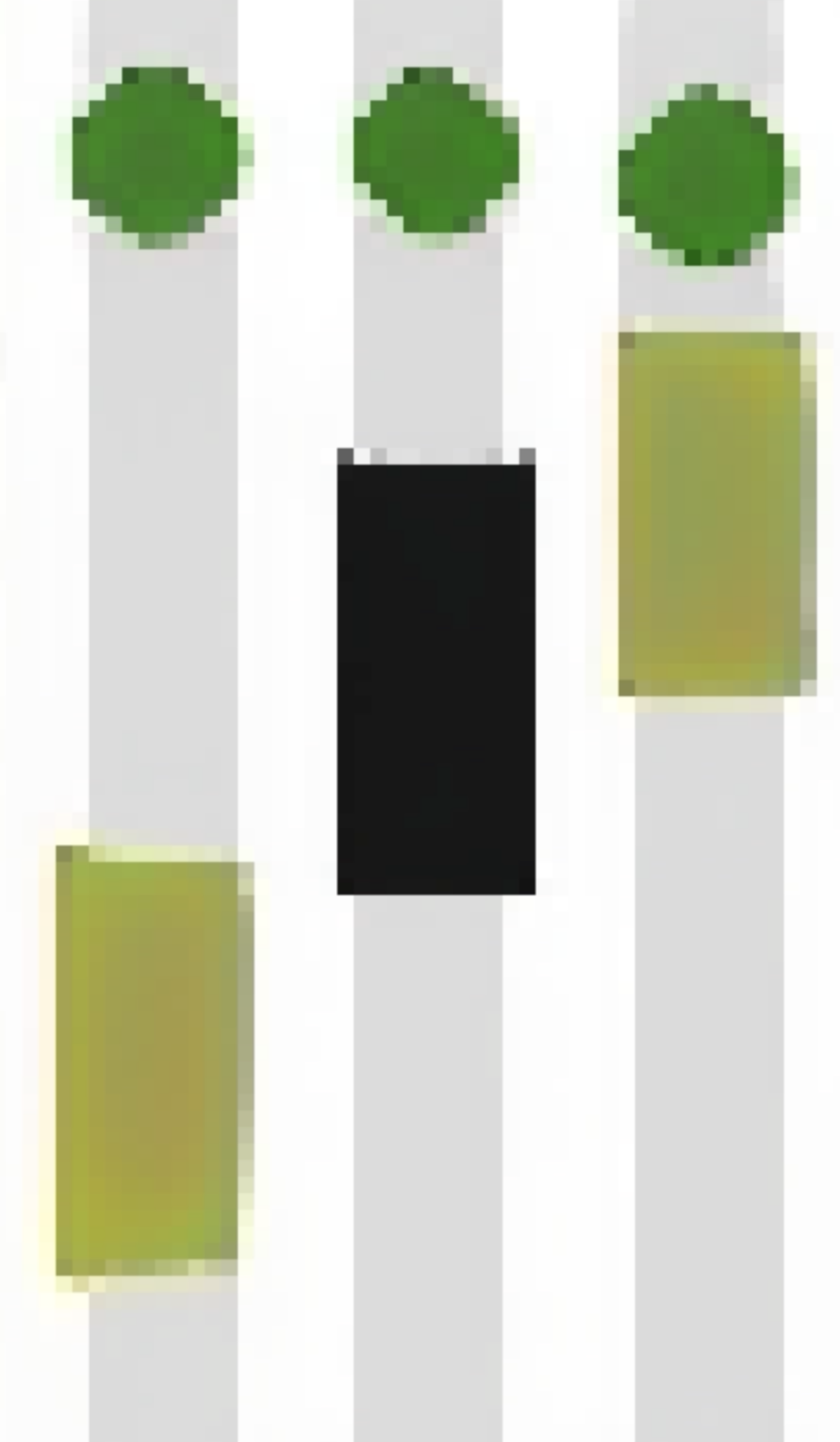}
        \put(21,-12){\small (b)}
    \end{overpic}
    \hspace{0.12em}
    \begin{overpic}[width=0.14\textwidth]{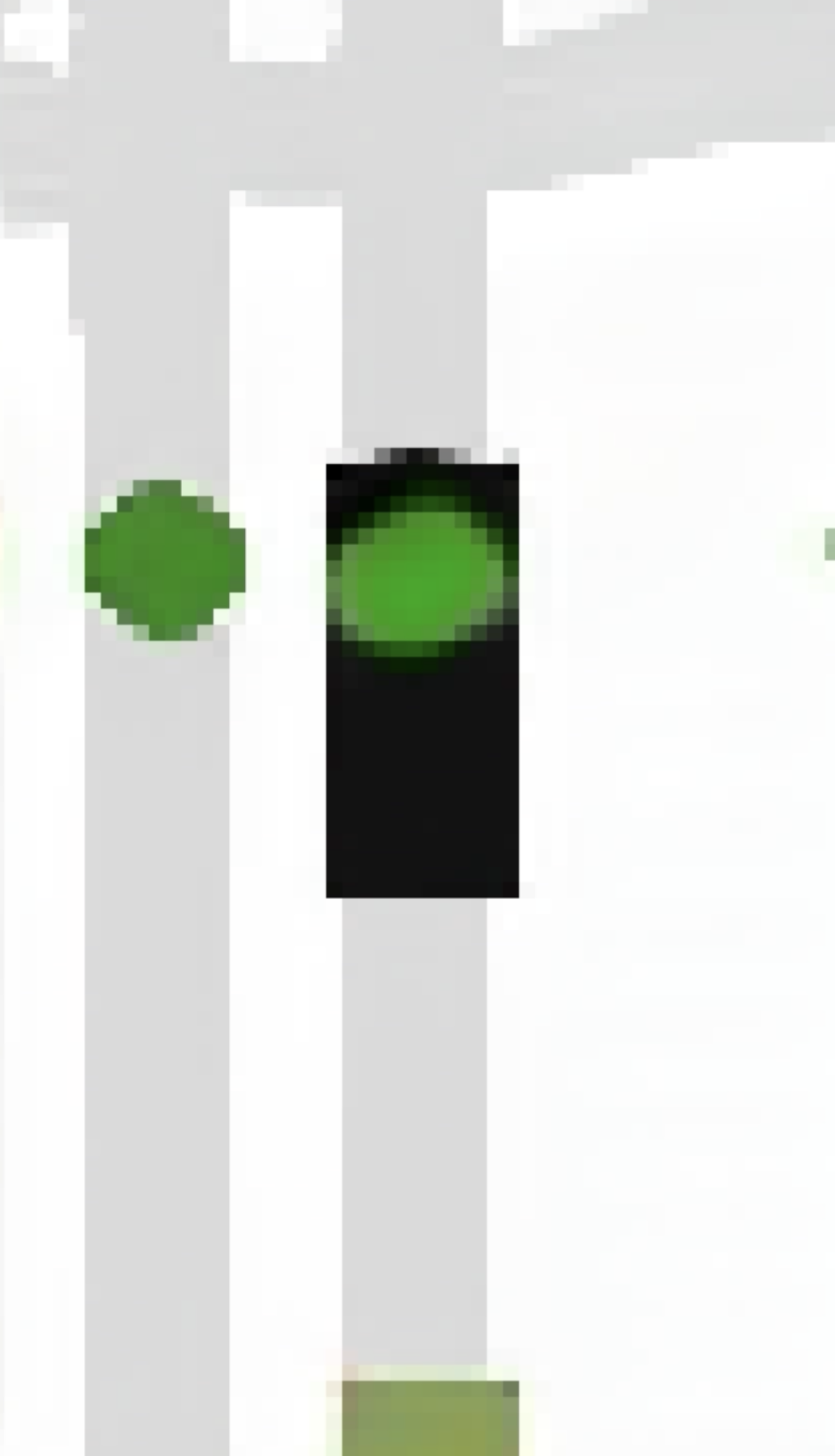}
        \put(21,-12){\small (c)}
    \end{overpic}
    \hspace{0.12em}
    \begin{overpic}[width=0.14\textwidth]{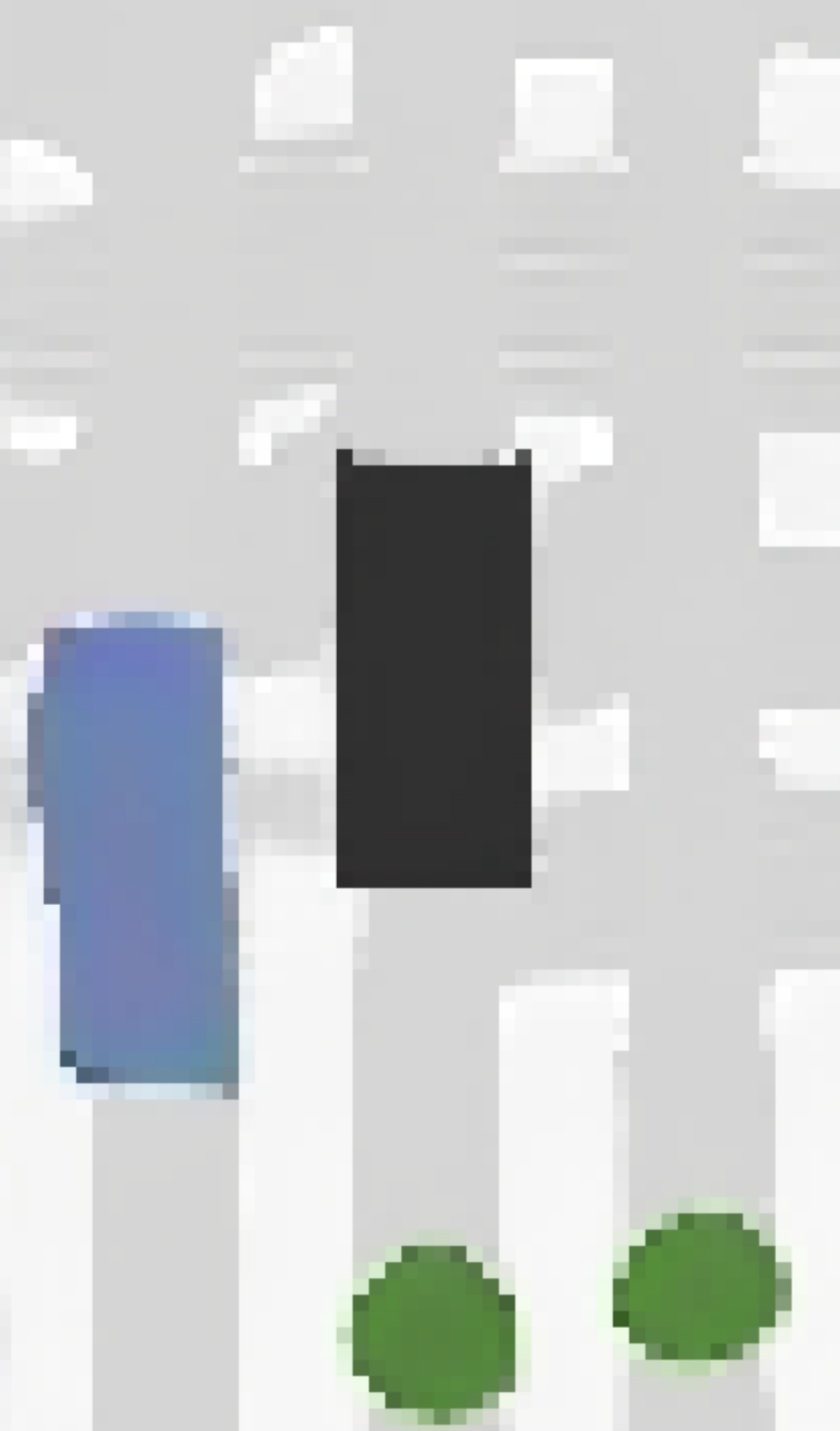}
        \put(21,-12){\small (d)}
    \end{overpic}
    \hspace{0.12em}
    \begin{overpic}[width=0.14\textwidth]{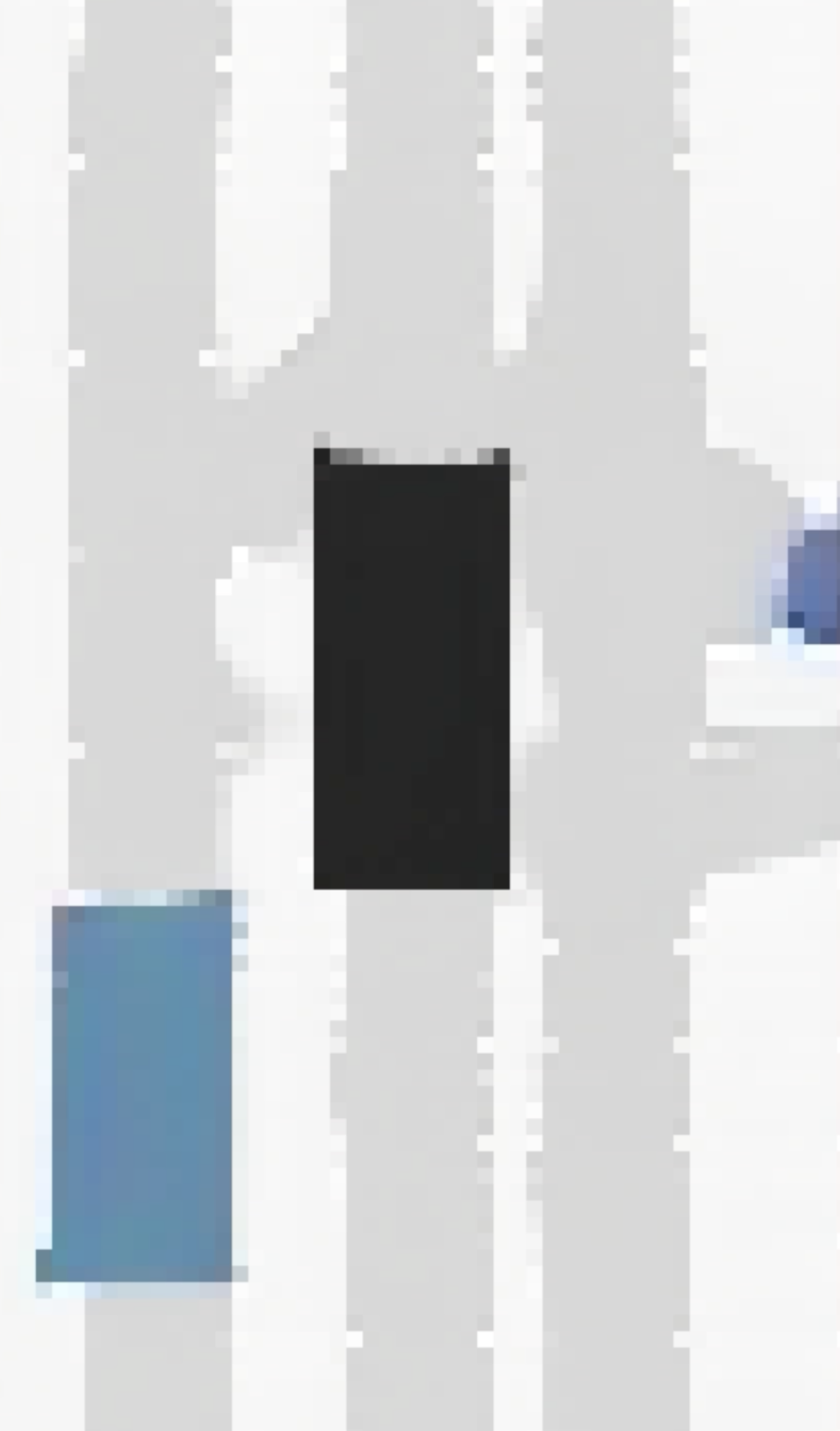}
        \put(21,-12){\small (e)}
    \end{overpic}
    \hspace{0.12em}
    \begin{overpic}[width=0.14\textwidth]{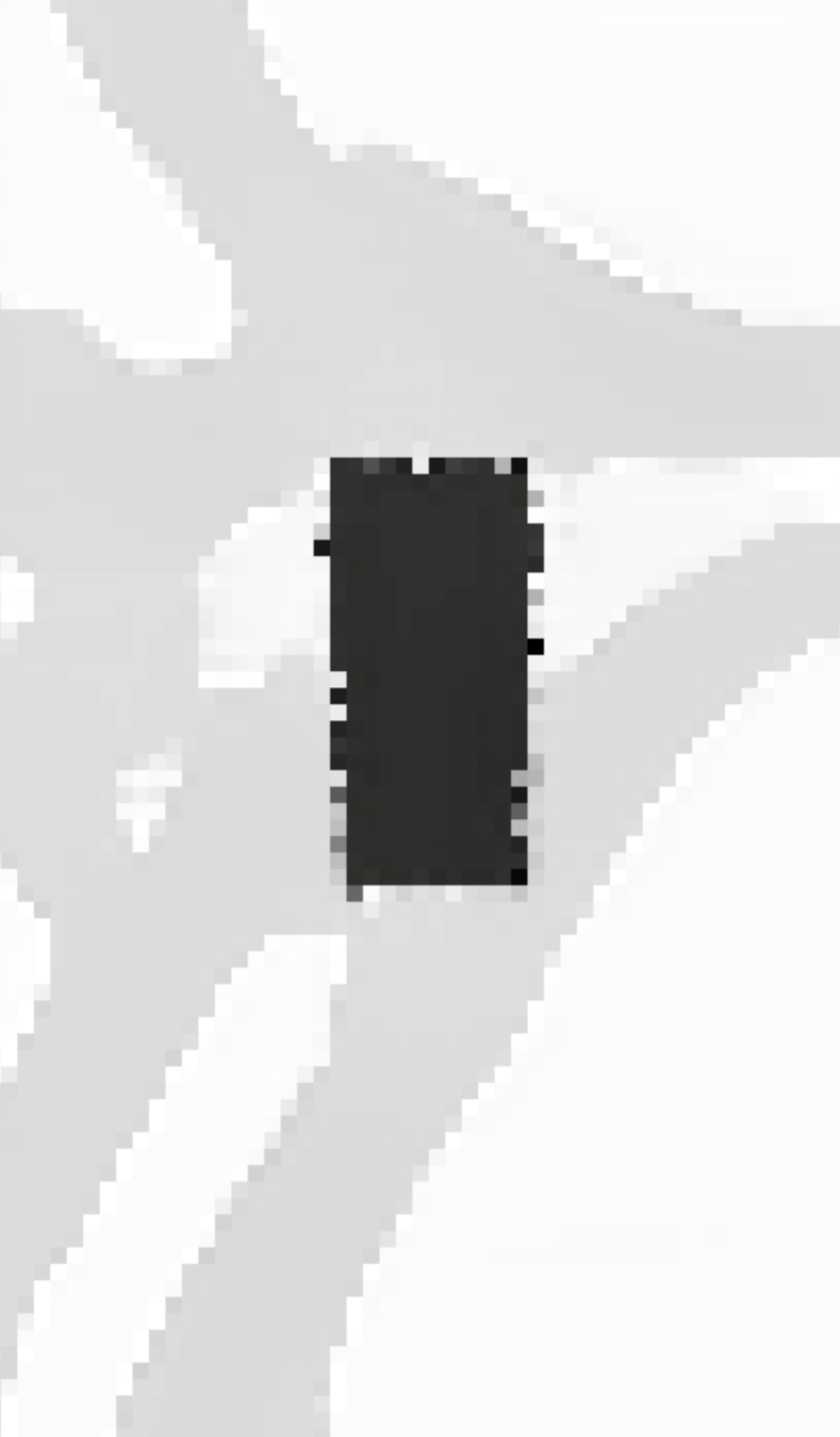}
        \put(21,-12){\small (f)}
    \end{overpic}
    \vspace{4ex}
    \caption{Examples from generated videos of traffic scenes. The agents are observed waiting at a stoplight (a), passing a green light (b,c), passing crossroads (d,e), and turning (f). In the right side of (e), notice the head of a vehicle yielding to the main traffic direction.}
    \label{fig:frame_examples}
\end{figure}

\paragraph{Qualitative results}
We provide a selection of video clips of the low-resolution BEV occupancy grid scenes rasterized from the Waymo dataset (hereafter referred to as "real") as well as some examples extracted from the generated videos at \url{https://www.youtube.com/watch?v=4Awa5zf26io} First, we observe that the generated map lines are comparable to those in the real videos in size and distancing, and the same applies to road users and traffic lights. With respect to those lines, the agents drive well-aligned with the center lane, and their dynamics appear mostly realistic. While agent identity is mostly retained in time, in some occasions a non-ego agent will morph length-wise and seem to split or disappear on the side. However, this is almost only observed near a crossing, which suggests the model is trying to model an agent entering or leaving the scene by turning at a crossroads. In rare cases, two agents merge into a single elongated form, which then rapidly resizes to normal vehicle length. On the other hand, sideways merging has never been observed. Crossings are featured as lane reticulations, relatively similar to the complex crossing patterns featured in the real videos (see Figure~\ref{fig:example} for a real scene and Figure~\ref{fig:frame_examples} for a generated one). Turning is featured as a morphing of the map in correspondence of a crossing, typically after a traffic light, after which a regular road layout resumes. 
In addition to this, the generated videos feature diverse and accurate behavior of the ego agent with respect to the traffic lights. While yellow lights are rarely observed, red and green lights are mostly realistic in size and color, and we find that the ego agent proceeds when the light is green, but slows down and stops at the red light. The model also generates dynamic color changes, and the agents behave accordingly. For example, when the light turns green the ego car which was waiting at the stoplight is observed to resume its motion.
Figure~\ref{fig:frame_examples} shows some examples of typically observed situations, such as agents waiting at a stoplight (a), passing a green light (b,c), passing crossroads (d,e) and turning (f). In all cases, the agents appear to keep a safety distance. 
These qualitative observations are quantitatively investigated in the following paragraphs.

\begin{figure}[t]
    \centering
    \includegraphics[width=0.46\textwidth]{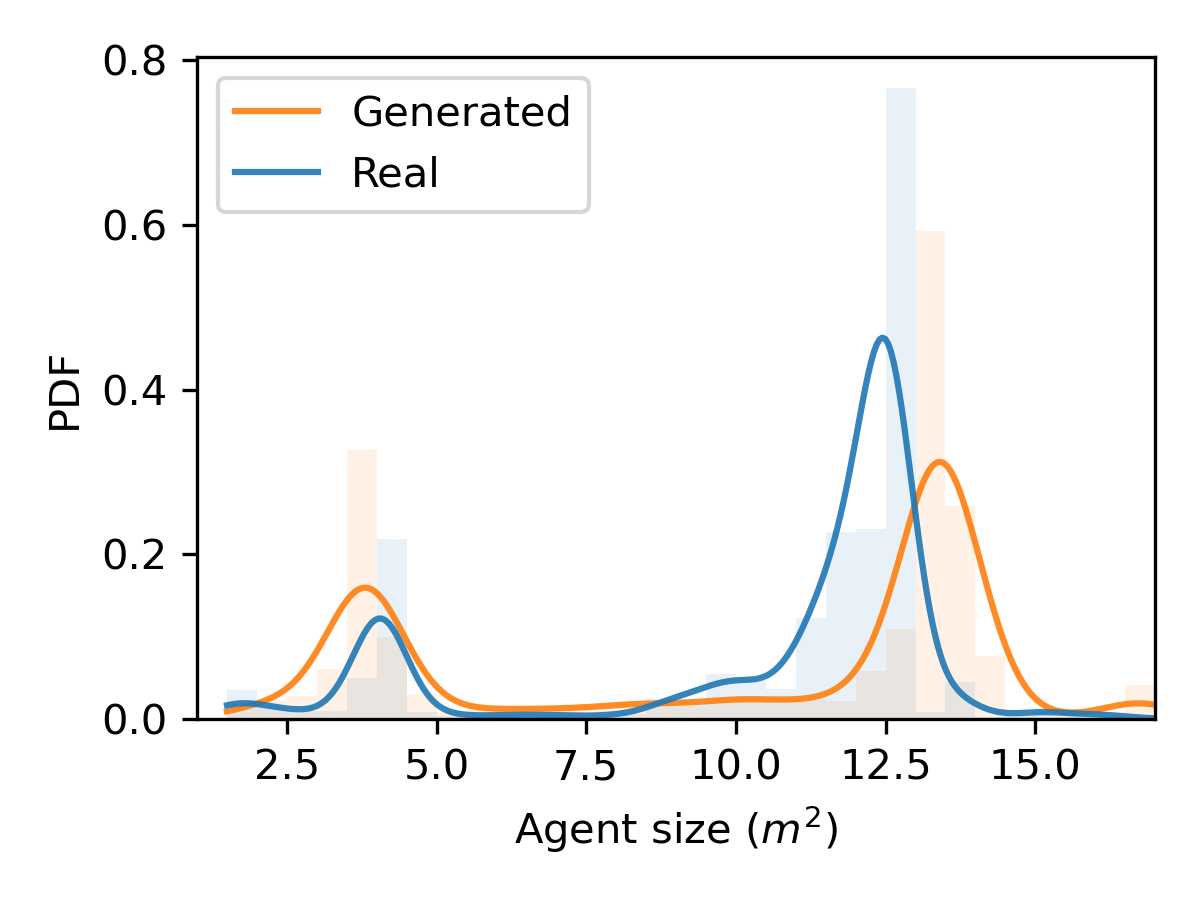}
    \includegraphics[width=0.46\textwidth]{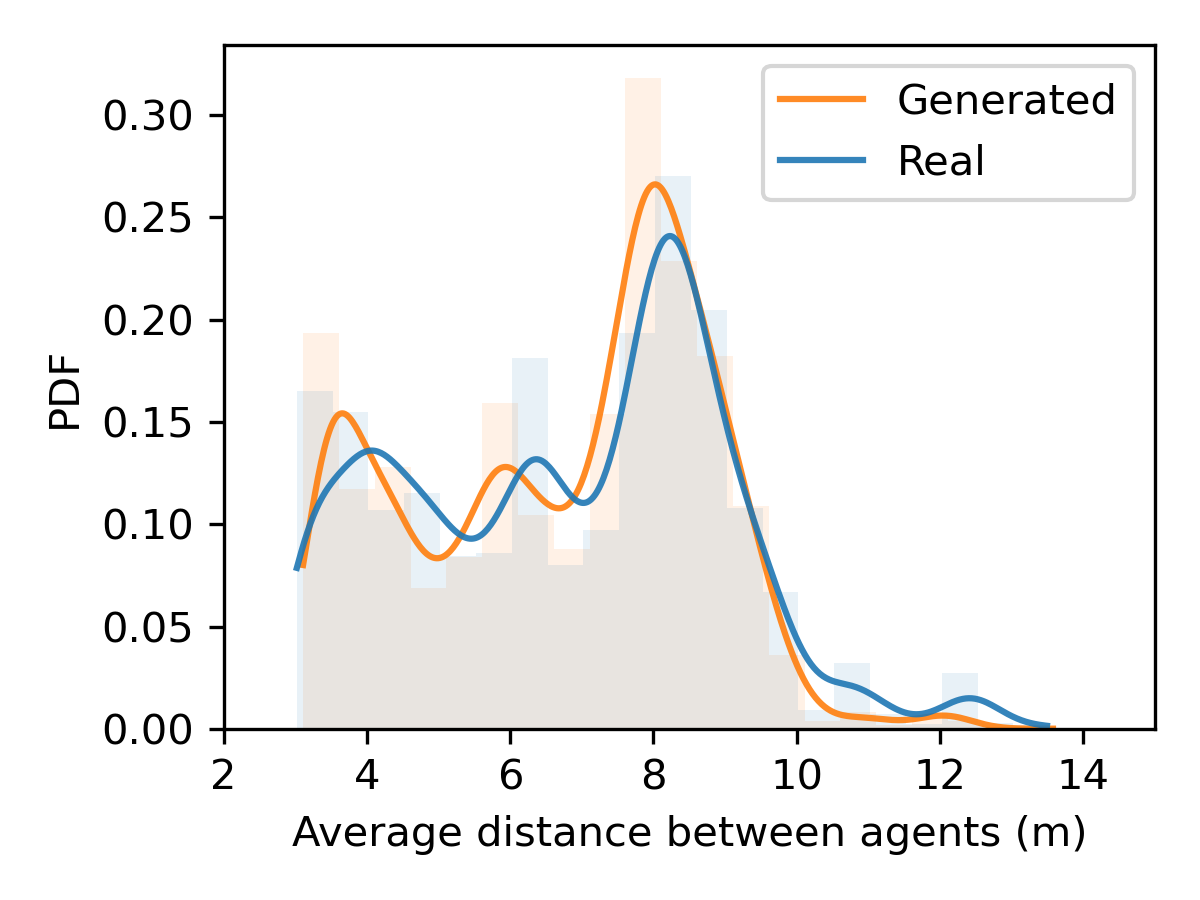}
    \includegraphics[width=0.46\textwidth]{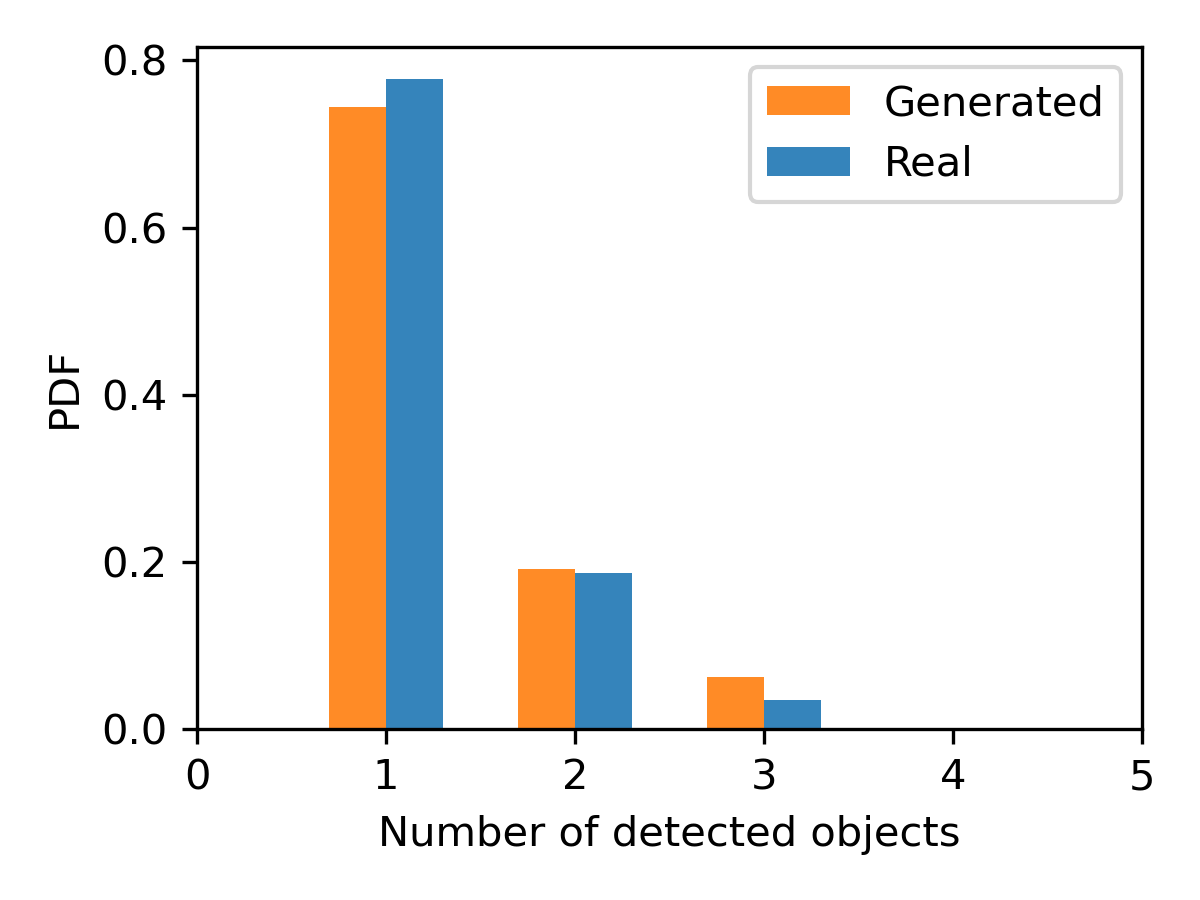}
    \includegraphics[width=0.46\textwidth]{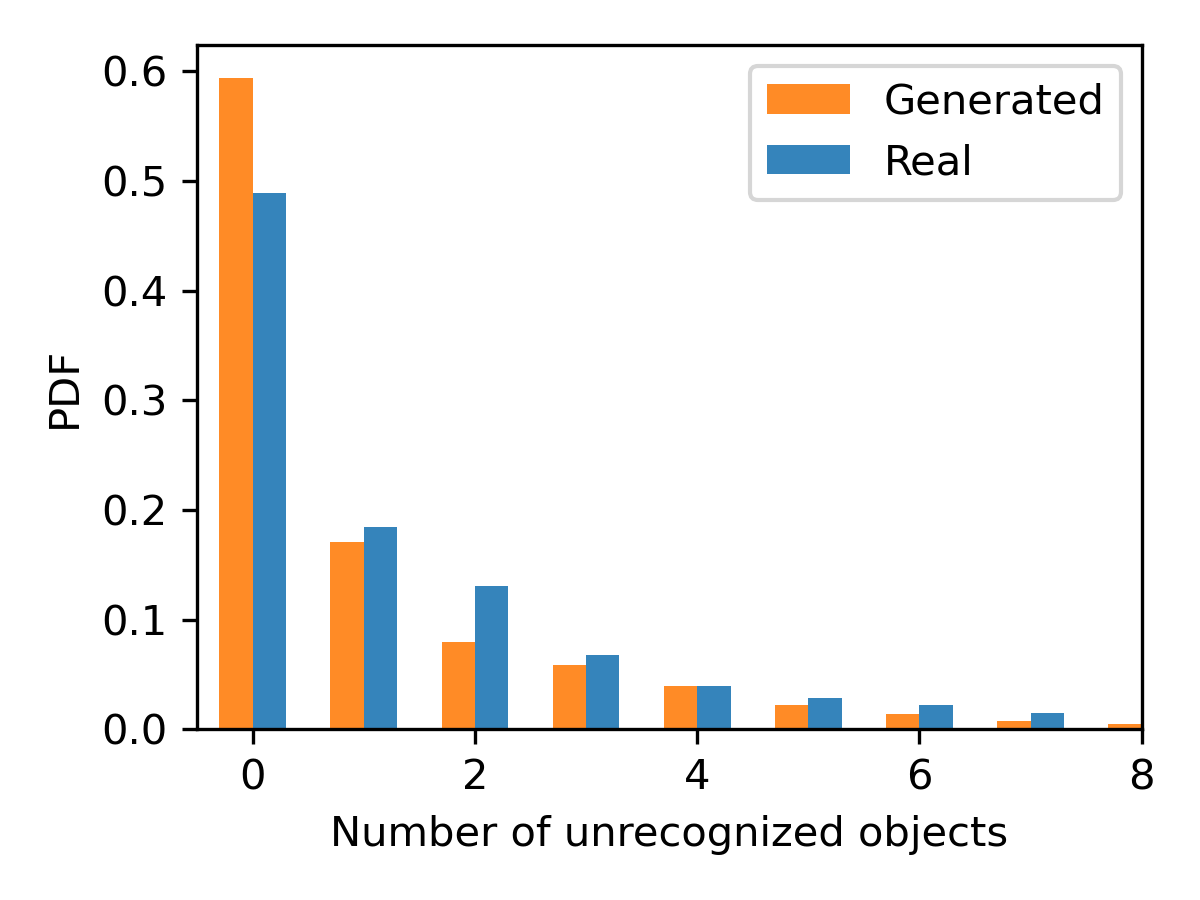}
    \caption{Comparison between the probability density distributions (PDF) of static parameters in the real and generated videos. Top left: agent sizes. Top right: minimum center-to-center distances in scenes with more than two agents. Bottom left: traffic density, defined as the number of identified objects per frame. Bottom right: number of unrecognized objects per frame.}
    \label{fig:static_comparisons}
\end{figure}

\paragraph{Quantitative results: frame by frame}
We study the distribution alignment for different geometrical parameters extracted from the videos, comparing real and generated videos. We apply the same object detection pipeline to 80 randomly generated videos and 120 real videos from the Waymo dataset. The distribution of object sizes (top-left plot in Figure~\ref{fig:static_comparisons}) is bimodal, with the peak around 4$\text{m}^2$ corresponding to traffic lights, and the peaks around 12-14$\text{m}^2$ corresponding to the road users, in both cases compatible with the average size of a car. While both peaks are comparable in intensity and position, the generated vehicles are on average 10\% larger than the real ones. In the top-right corner, the average distance between agents in traffic is remarkably similar for generated and real scenes. In Figure~\ref{fig:static_comparisons} bottom left, it is shown that the traffic density is also comparable, with the majority of frames in both cases containing only the ego agent or one more road user, and no more than 3 full size vehicles observed per scene, which is due to the relatively small visual surroundings. The similar distribution of the amount of unidentified object reinforces the qualitative observation above regarding the realism of the frames (see bottom right plot).

\begin{figure}[bt]
    \centering

    \includegraphics[width=0.32\textwidth]{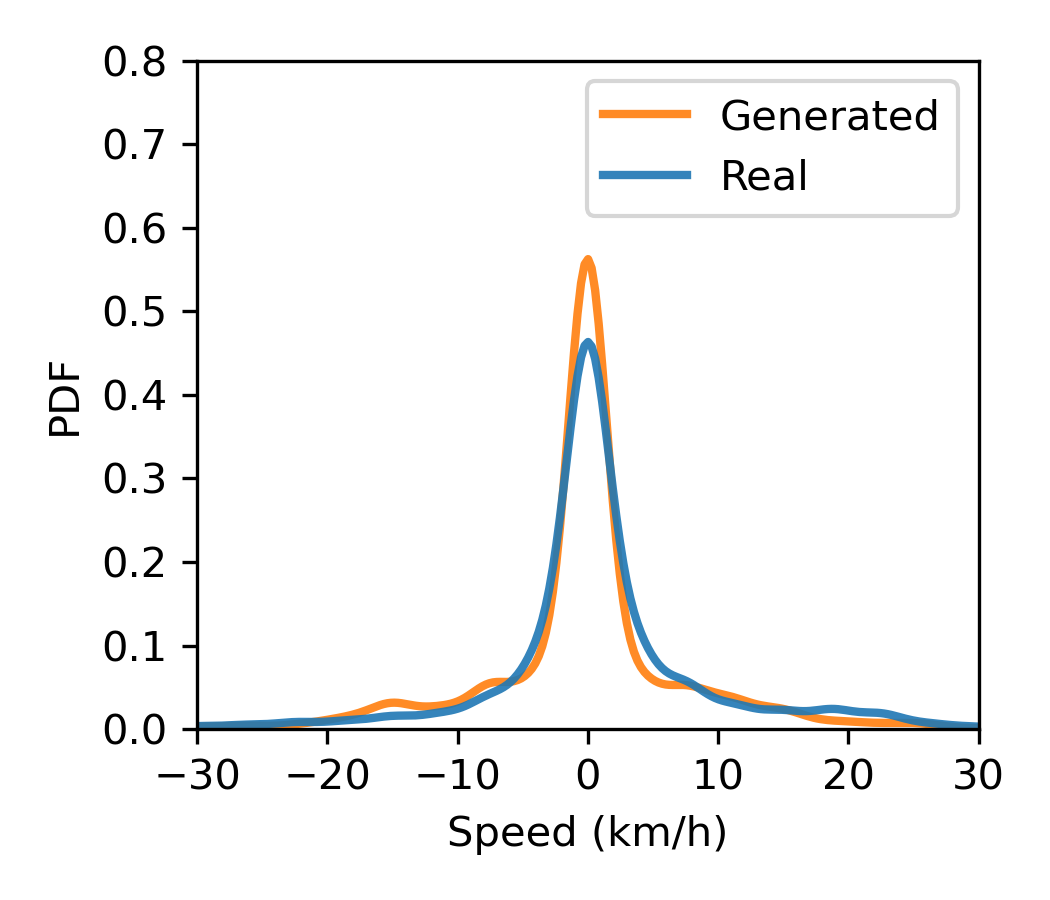}
    \includegraphics[width=0.32\textwidth]{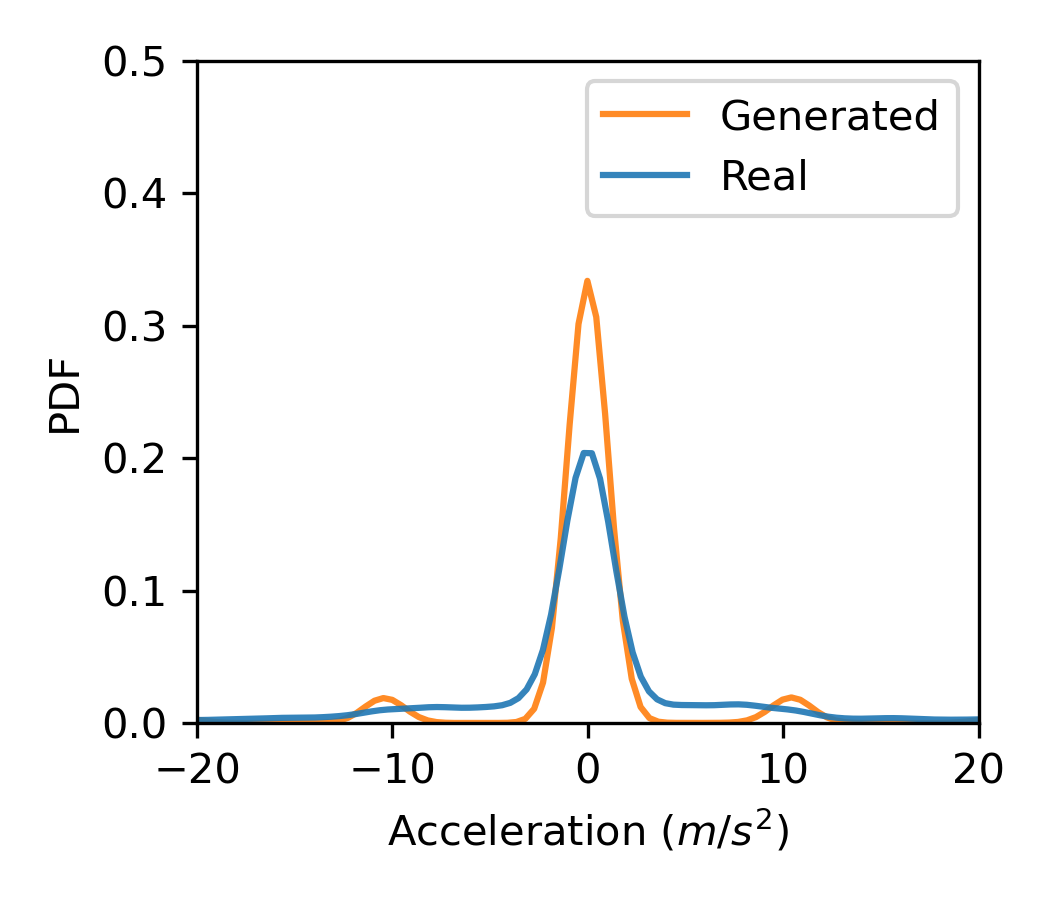}
    \includegraphics[width=0.32\textwidth]{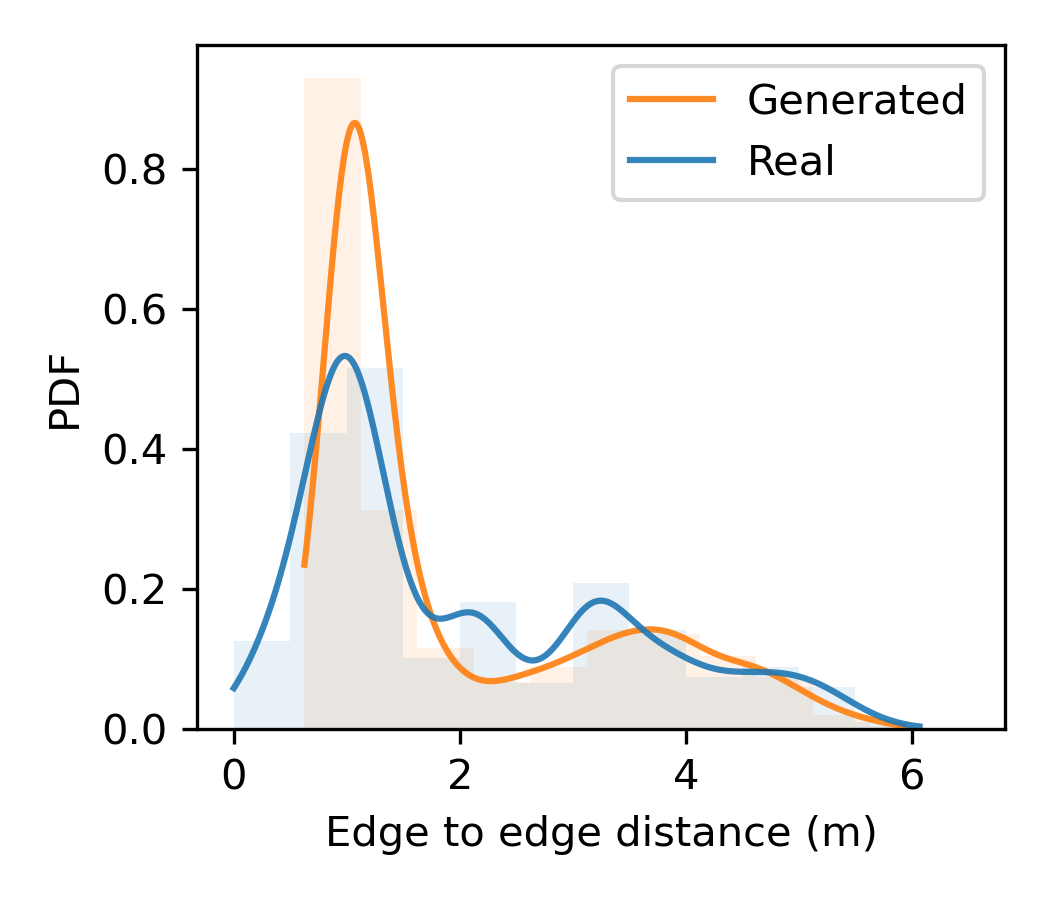}

    \caption{Comparison between the probability density distributions (PDF) of the dynamic parameters
    for road users.}
    \label{fig:dynamics}
\end{figure}

\paragraph{Quantitative results: dynamics}
The statistical distributions of speeds and accelerations of agents relatively to the ego vehicle, shown in Figure~\ref{fig:dynamics}, confirm that the model is successful at capturing and reproducing the dynamics of the traffic scenes. This matches what we qualitatively observed watching the videos. Positive speeds and accelerations are due to road users overtaking the ego vehicle, and similarly negative speeds are for agents moving in the opposite direction or being overtaken by the ego vehicle. The sizable peak around 0 is partially due to co-moving traffic and partly to the fact that many scenes feature only the ego agent. We also report the minimum edge-to-edge distance between agents in a frame, which confirms that road users maintain appropriate separation in the traffic.

\begin{figure}[tb]
    \centering
    \includegraphics[width=0.47\textwidth]{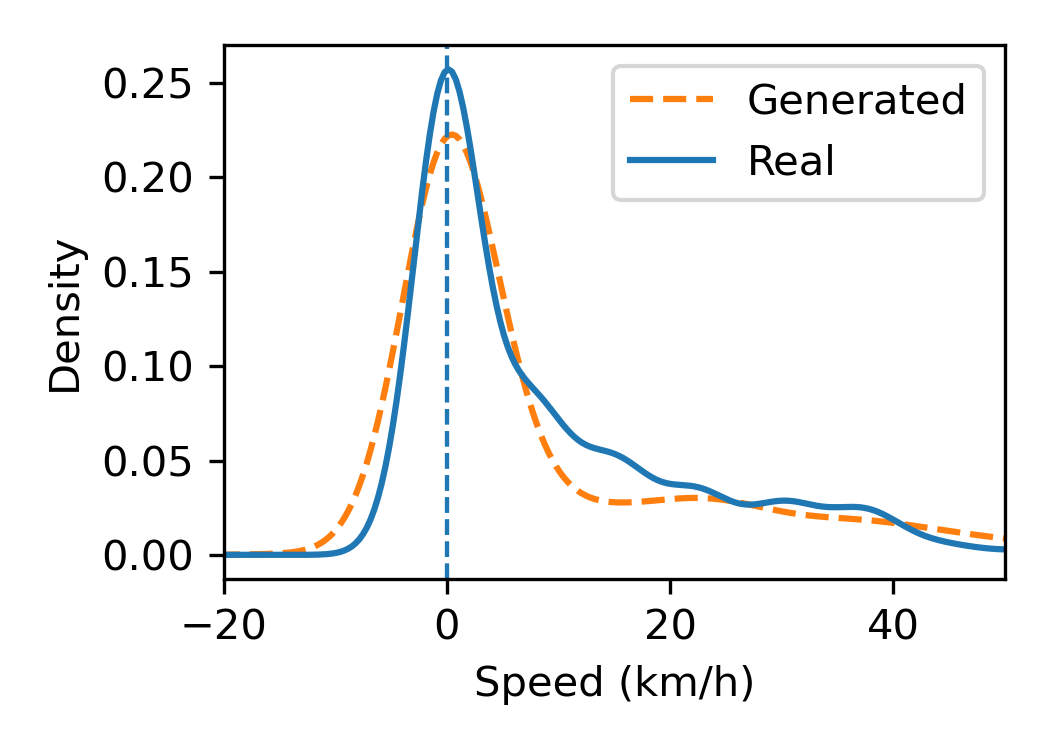}
    \includegraphics[width=0.47\textwidth]{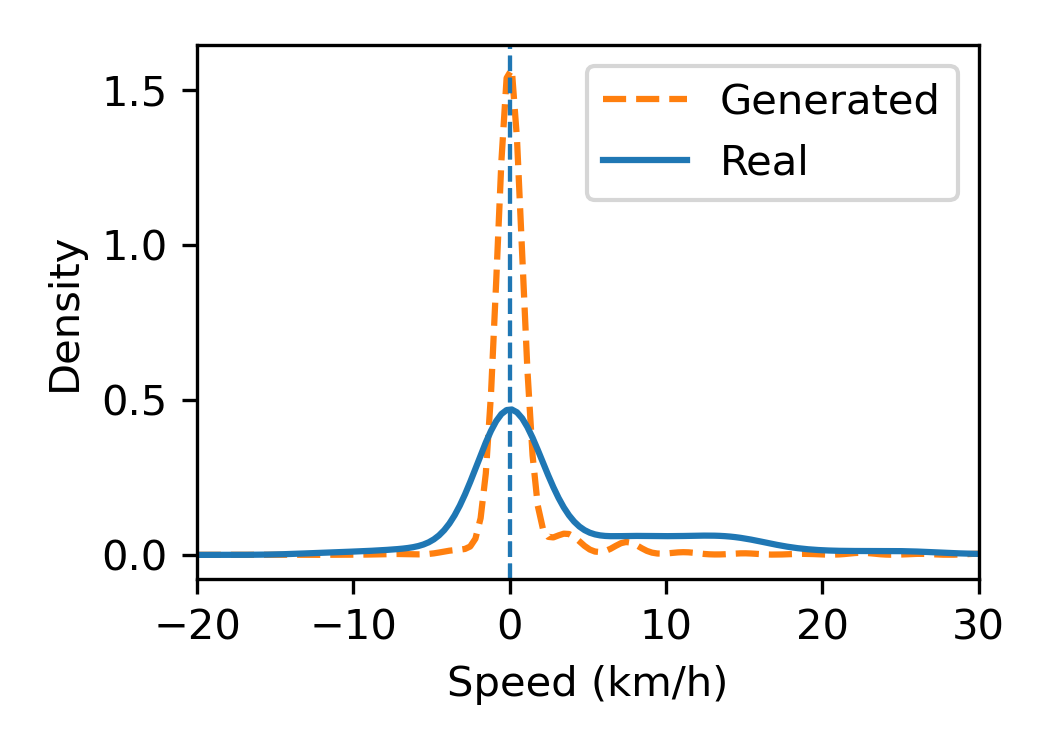}
    \caption{Distribution of ego vehicle speed when approaching or passing a green traffic light (left) and when approaching or waiting at a red light (right).}
    \label{fig:traffic_lights}
\end{figure}

\paragraph{Quantitative results: traffic lights} We present a quantitative measure of the interaction of the ego vehicle with the traffic signals. As already observed qualitatively, the vehicle is likely to pass a green light either at low speed, or at a normal traffic speed (see Figure~\ref{fig:traffic_lights} on the left). The difference between real and generated data is almost negligible in this plot. For the red lights, shown in Figure~\ref{fig:traffic_lights} on the right, 0\,km/h is the most featured speed, confirming that the model has learned to stop at the red light.

\section{Conclusions}
In this work, we have introduced a new trajectory proposal pipeline based on a video generative model. Our approach leverages the benefits of both visually-grounded approaches, which excel at modeling spatial interactions between agents and infrastructures, and of generative models, which excel at modeling complex and diverse distributions. In particular, we built a dataset of low-resolution BEV occupancy grids videos of traffic scenes rasterized from abstract data and used these as input for a videoGan model. Then, we extracted trajectory data from the generated videos using single-frame object detection and frame-to-frame object matching. We analyzed the distribution alignment of spatial and dynamic parameters such as inter-agent distances and relative speeds with respect to the real data, along with similar analysis of the interaction of road users with the dynamic traffic lights. These experiments show that our model learned to generate realistic and safe traffic scenes and trajectories, notably including dynamic interaction with dynamic traffic signals. We achieve inference speeds of 20\,ms, in principle suitable for real-time applications.
Future work should investigate modifications or replacements of the generative architecture and study its potential in handling larger and more complex scenes to capture behaviors such as yielding for another road user or pedestrian crossing the road.

\section{Acknowledgments}
This project has been supported by the German Federal Ministry for Economic Affairs and Climate Action within the project "NXT GEN AI METHODS – Generative Methoden für Perzeption, Prädiktion und Planung" (Grant no.\ 19A23014Q). The authors gratefully acknowledge the Gauss Centre for Supercomputing e.V. \url{https://www.gauss-centre.eu} for funding this project by providing computing time on the GCS Supercomputer JUWELS at Jülich Supercomputing Centre (JSC).

\bibliography{references_zotero}

\begin{thebibliography}{44}
\providecommand{\natexlab}[1]{#1}
\providecommand{\url}[1]{\texttt{#1}}
\expandafter\ifx\csname urlstyle\endcsname\relax
  \providecommand{\doi}[1]{doi: #1}\else
  \providecommand{\doi}{doi: \begingroup \urlstyle{rm}\Url}\fi

\bibitem[Agro et~al.(2023)Agro, Sykora, Casas, and Urtasun]{agro_implicit_2023}
Ben Agro, Quinlan Sykora, Sergio Casas, and Raquel Urtasun.
\newblock Implicit {{Occupancy Flow Fields}} for {{Perception}} and {{Prediction}} in {{Self-Driving}}.
\newblock In \emph{Proceedings of the {{IEEE}}/{{CVF Conference}} on {{Computer Vision}} and {{Pattern Recognition}}}, pages 1379--1388, 2023.

\bibitem[Alahi et~al.(2016)Alahi, Goel, Ramanathan, Robicquet, {Fei-Fei}, and Savarese]{alahi_social_2016}
Alexandre Alahi, Kratarth Goel, Vignesh Ramanathan, Alexandre Robicquet, Li~{Fei-Fei}, and Silvio Savarese.
\newblock Social {{LSTM}}: {{Human Trajectory Prediction}} in {{Crowded Spaces}}.
\newblock In \emph{2016 {{IEEE Conference}} on {{Computer Vision}} and {{Pattern Recognition}} ({{CVPR}})}, pages 961--971, June 2016.
\newblock \doi{10.1109/CVPR.2016.110}.

\bibitem[Ashwin and Naveen~Raj(2023)]{ashwin_deep_2023}
S.~H. Ashwin and Rashmi Naveen~Raj.
\newblock Deep reinforcement learning for autonomous vehicles: Lane keep and overtaking scenarios with collision avoidance.
\newblock \emph{Int. j. inf. tecnol.}, 15\penalty0 (7):\penalty0 3541--3553, October 2023.
\newblock ISSN 2511-2112.
\newblock \doi{10.1007/s41870-023-01412-6}.

\bibitem[Bharilya and Kumar(2023)]{bharilya_machine_2023}
Vibha Bharilya and Neetesh Kumar.
\newblock Machine {{Learning}} for {{Autonomous Vehicle}}'s {{Trajectory Prediction}}: {{A}} comprehensive survey, {{Challenges}}, and {{Future Research Directions}}, July 2023.

\bibitem[Brooks et~al.(2022)Brooks, Hellsten, Aittala, Wang, Aila, Lehtinen, Liu, Efros, and Karras]{brooks_generating_2022}
Tim Brooks, Janne Hellsten, Miika Aittala, Ting-Chun Wang, Timo Aila, Jaakko Lehtinen, Ming-Yu Liu, Alexei~A Efros, and Tero Karras.
\newblock Generating {{Long Videos}} of {{Dynamic Scenes}}.
\newblock In \emph{36th {{Conference}} on {{Neural Information Processing Systems}} ({{NeurIPS}} 2022)}, 2022.

\bibitem[Casas et~al.(2021)Casas, Sadat, and Urtasun]{casas_mp3_2021}
Sergio Casas, Abbas Sadat, and Raquel Urtasun.
\newblock {{MP3}}: {{A Unified Model To Map}}, {{Perceive}}, {{Predict}} and {{Plan}}.
\newblock In \emph{Proceedings of the {{IEEE}}/{{CVF Conference}} on {{Computer Vision}} and {{Pattern Recognition}}}, pages 14403--14412, 2021.

\bibitem[Chen et~al.(2025)Chen, Huang, Wang, and Chen]{chen_trajectory_2025}
Xin Chen, Chengrui Huang, Chenhao Wang, and Lisi Chen.
\newblock Trajectory generation: A survey on methods and techniques.
\newblock \emph{Geoinformatica}, April 2025.
\newblock ISSN 1573-7624.
\newblock \doi{10.1007/s10707-025-00545-z}.

\bibitem[Cui et~al.(2021)Cui, Casas, Sadat, Liao, and Urtasun]{cui_lookout_2021}
Alexander Cui, Sergio Casas, Abbas Sadat, Renjie Liao, and Raquel Urtasun.
\newblock {{LookOut}}: {{Diverse Multi-Future Prediction}} and {{Planning}} for {{Self-Driving}}.
\newblock In \emph{Proceedings of the {{IEEE}}/{{CVF International Conference}} on {{Computer Vision}}}, pages 16107--16116, 2021.

\bibitem[Fang et~al.(2020)Fang, Jiang, Shi, and Zhou]{fang_tpnet_2020}
Liangji Fang, Qinhong Jiang, Jianping Shi, and Bolei Zhou.
\newblock {{TPNet}}: {{Trajectory Proposal Network}} for {{Motion Prediction}}.
\newblock In \emph{Proceedings of the {{IEEE}}/{{CVF Conference}} on {{Computer Vision}} and {{Pattern Recognition}}}, pages 6797--6806, 2020.

\bibitem[Gilles et~al.(2021)Gilles, Sabatini, Tsishkou, Stanciulescu, and Moutarde]{gilles_home_2021}
Thomas Gilles, Stefano Sabatini, Dzmitry Tsishkou, Bogdan Stanciulescu, and Fabien Moutarde.
\newblock {{HOME}}: {{Heatmap Output}} for future {{Motion Estimation}}.
\newblock In \emph{{{IEEE Int}}. {{Intell}}. {{Transp}}. {{Syst}}. {{Conf}}.}, pages 500--507, 2021.
\newblock \doi{10.1109/ITSC48978.2021.9564944}.

\bibitem[Grigorescu et~al.(2020)Grigorescu, Trasnea, Cocias, and Macesanu]{grigorescu_survey_2020}
Sorin Grigorescu, Bogdan Trasnea, Tiberiu Cocias, and Gigel Macesanu.
\newblock A {{Survey}} of {{Deep Learning Techniques}} for {{Autonomous Driving}}.
\newblock \emph{Journal of Field Robotics}, 37\penalty0 (3):\penalty0 362--386, April 2020.
\newblock ISSN 1556-4959, 1556-4967.
\newblock \doi{10.1002/rob.21918}.

\bibitem[Gu et~al.(2022)Gu, Chen, Li, Lin, Rao, Zhou, and Lu]{gu_stochastic_2022}
Tianpei Gu, Guangyi Chen, Junlong Li, Chunze Lin, Yongming Rao, Jie Zhou, and Jiwen Lu.
\newblock Stochastic {{Trajectory Prediction}} via {{Motion Indeterminacy Diffusion}}.
\newblock In \emph{Proceedings of the {{IEEE}}/{{CVF Conference}} on {{Computer Vision}} and {{Pattern Recognition}}}, pages 17113--17122, 2022.

\bibitem[Gupta et~al.(2018)Gupta, Johnson, {Fei-Fei}, Savarese, and Alahi]{gupta_social_2018}
Agrim Gupta, Justin Johnson, Li~{Fei-Fei}, Silvio Savarese, and Alexandre Alahi.
\newblock Social {{GAN}}: {{Socially Acceptable Trajectories With Generative Adversarial Networks}}.
\newblock In \emph{Proceedings of the {{IEEE Conference}} on {{Computer Vision}} and {{Pattern Recognition}}}, pages 2255--2264, 2018.

\bibitem[Hagedorn et~al.(2024)Hagedorn, Hallgarten, Stoll, and Condurache]{hagedorn_integration_2024}
Steffen Hagedorn, Marcel Hallgarten, Martin Stoll, and Alexandru~Paul Condurache.
\newblock The {{Integration}} of {{Prediction}} and {{Planning}} in {{Deep Learning Automated Driving Systems}}: {{A Review}}.
\newblock \emph{IEEE Transactions on Intelligent Vehicles}, pages 1--17, 2024.
\newblock ISSN 2379-8904.
\newblock \doi{10.1109/TIV.2024.3459071}.

\bibitem[Horgan et~al.(2015)Horgan, Hughes, McDonald, and Yogamani]{horgan_vision-based_2015}
Jonathan Horgan, Ciar{\'a}n Hughes, John McDonald, and Senthil Yogamani.
\newblock Vision-{{Based Driver Assistance Systems}}: {{Survey}}, {{Taxonomy}} and {{Advances}}.
\newblock In \emph{Proceedings of the 2015 {{IEEE}} 18th {{International Conference}} on {{Intelligent Transportation Systems}}}, {{ITSC}} '15, pages 2032--2039, USA, September 2015. IEEE Computer Society.
\newblock ISBN 978-1-4673-6596-3.
\newblock \doi{10.1109/ITSC.2015.329}.

\bibitem[Hu et~al.(2021)Hu, Murez, Mohan, Dudas, Hawke, Badrinarayanan, Cipolla, and Kendall]{hu_fiery_2021}
Anthony Hu, Zak Murez, Nikhil Mohan, Sofia Dudas, Jeffrey Hawke, Vijay Badrinarayanan, Roberto Cipolla, and Alex Kendall.
\newblock {{FIERY}}: {{Future Instance Prediction}} in {{Bird}}'s-{{Eye View}} from {{Surround Monocular Cameras}}.
\newblock In \emph{2021 {{IEEE}}/{{CVF International Conference}} on {{Computer Vision}} ({{ICCV}})}, pages 15253--15262, Montreal, QC, Canada, October 2021. IEEE.
\newblock ISBN 978-1-66542-812-5.
\newblock \doi{10.1109/ICCV48922.2021.01499}.

\bibitem[Hu et~al.(2022)Hu, Chen, Wu, Li, Yan, and Tao]{hu_st-p3_2022}
Shengchao Hu, Li~Chen, Penghao Wu, Hongyang Li, Junchi Yan, and Dacheng Tao.
\newblock {{ST-P3}}: {{End-to-End Vision-Based Autonomous Driving}} via~{{Spatial-Temporal Feature Learning}}.
\newblock In Shai Avidan, Gabriel Brostow, Moustapha Ciss{\'e}, Giovanni~Maria Farinella, and Tal Hassner, editors, \emph{Computer {{Vision}} -- {{ECCV}} 2022}, pages 533--549, Cham, 2022. Springer Nature Switzerland.
\newblock ISBN 978-3-031-19839-7.
\newblock \doi{10.1007/978-3-031-19839-7_31}.

\bibitem[Huang et~al.(2023{\natexlab{a}})Huang, Zhuo, Xiong, Lu, and Tian]{huang_review_2023}
Renbo Huang, Guirong Zhuo, Lu~Xiong, Shouyi Lu, and Wei Tian.
\newblock A {{Review}} of {{Deep Learning-Based Vehicle Motion Prediction}} for {{Autonomous Driving}}.
\newblock \emph{Sustainability}, 15\penalty0 (20):\penalty0 14716, October 2023{\natexlab{a}}.
\newblock ISSN 2071-1050.
\newblock \doi{10.3390/su152014716}.

\bibitem[Huang et~al.(2023{\natexlab{b}})Huang, Xue, Pagnucco, Salim, and Song]{huang_multimodal_2023}
Renhao Huang, Hao Xue, Maurice Pagnucco, Flora Salim, and Yang Song.
\newblock Multimodal {{Trajectory Prediction}}: {{A Survey}}, February 2023{\natexlab{b}}.

\bibitem[Jia et~al.(2023)Jia, Wu, Chen, Liu, Li, and Yan]{jia_hdgt_2023}
Xiaosong Jia, Penghao Wu, Li~Chen, Yu~Liu, Hongyang Li, and Junchi Yan.
\newblock {{HDGT}}: {{Heterogeneous Driving Graph Transformer}} for {{Multi-Agent Trajectory Prediction}} via {{Scene Encoding}}.
\newblock \emph{IEEE Trans. Pattern Anal. Mach. Intell.}, 45\penalty0 (11):\penalty0 13860--13875, November 2023.
\newblock ISSN 0162-8828, 2160-9292, 1939-3539.
\newblock \doi{10.1109/TPAMI.2023.3298301}.

\bibitem[Jiang et~al.(2023)Jiang, Cornman, Park, Sapp, Zhou, and Anguelov]{jiang_motiondiffuser_2023}
Chiyu~``Max'' Jiang, Andre Cornman, Cheolho Park, Benjamin Sapp, Yin Zhou, and Dragomir Anguelov.
\newblock {{MotionDiffuser}}: {{Controllable Multi-Agent Motion Prediction Using Diffusion}}.
\newblock In \emph{Proceedings of the {{IEEE}}/{{CVF Conference}} on {{Computer Vision}} and {{Pattern Recognition}}}, pages 9644--9653, 2023.

\bibitem[Jiao et~al.(2024)Jiao, Wang, Liu, Zhan, Huang, and Zhu]{jiao_kinematics-aware_2024}
Ruochen Jiao, Yixuan Wang, Xiangguo Liu, Simon~Sinong Zhan, Chao Huang, and Qi~Zhu.
\newblock Kinematics-aware {{Trajectory Generation}} and {{Prediction}} with {{Latent Stochastic Differential Modeling}}.
\newblock In \emph{2024 {{IEEE}}/{{RSJ International Conference}} on {{Intelligent Robots}} and {{Systems}} ({{IROS}})}, pages 565--572, Abu Dhabi, United Arab Emirates, October 2024. IEEE.
\newblock ISBN 9798350377705.
\newblock \doi{10.1109/IROS58592.2024.10802438}.

\bibitem[Karnchanachari et~al.(2024)Karnchanachari, Geromichalos, Tan, Li, Eriksen, Yaghoubi, Mehdipour, Bernasconi, Fong, Guo, and Caesar]{karnchanachari_towards_2024}
Napat Karnchanachari, Dimitris Geromichalos, Kok~Seang Tan, Nanxiang Li, Christopher Eriksen, Shakiba Yaghoubi, Noushin Mehdipour, Gianmarco Bernasconi, Whye~Kit Fong, Yiluan Guo, and Holger Caesar.
\newblock Towards learning-based planning: {{The nuPlan}} benchmark for real-world autonomous driving.
\newblock In \emph{2024 {{IEEE International Conference}} on {{Robotics}} and {{Automation}} ({{ICRA}})}, pages 629--636, May 2024.
\newblock \doi{10.1109/ICRA57147.2024.10610077}.

\bibitem[Khan et~al.(2023)Khan, Sayed, Malik, Zia, Khan, Alkaabi, and Ignatious]{khan_level-5_2023}
Manzoor~Ahmed Khan, Hesham~El Sayed, Sumbal Malik, Talha Zia, Jalal Khan, Najla Alkaabi, and Henry Ignatious.
\newblock Level-5 {{Autonomous Driving}}---{{Are We There Yet}}? {{A Review}} of {{Research Literature}}.
\newblock \emph{ACM Comput. Surv.}, 55\penalty0 (2):\penalty0 1--38, February 2023.
\newblock ISSN 0360-0300, 1557-7341.
\newblock \doi{10.1145/3485767}.

\bibitem[Leon and Gavrilescu(2021)]{leon_review_2021}
Florin Leon and Marius Gavrilescu.
\newblock A {{Review}} of {{Tracking}} and {{Trajectory Prediction Methods}} for {{Autonomous Driving}}.
\newblock \emph{Mathematics}, 9\penalty0 (6):\penalty0 660, January 2021.
\newblock ISSN 2227-7390.
\newblock \doi{10.3390/math9060660}.

\bibitem[Li et~al.(2025)Li, Shao, Zhang, Wang, Brunswic, Zhou, Dong, Guo, Li, Chen, Wang, and Hao]{li_generative_2025}
Yinchuan Li, Xinyu Shao, Jianping Zhang, Haozhi Wang, Leo~Maxime Brunswic, Kaiwen Zhou, Jiqian Dong, Kaiyang Guo, Xiu Li, Zhitang Chen, Jun Wang, and Jianye Hao.
\newblock Generative {{Models}} in {{Decision Making}}: {{A Survey}}, March 2025.

\bibitem[Mahjourian et~al.(2022)Mahjourian, Kim, Chai, Tan, Sapp, and Anguelov]{mahjourian_occupancy_2022}
Reza Mahjourian, Jinkyu Kim, Yuning Chai, Mingxing Tan, Ben Sapp, and Dragomir Anguelov.
\newblock Occupancy {{Flow Fields}} for {{Motion Forecasting}} in {{Autonomous Driving}}.
\newblock \emph{IEEE Robot. Autom. Lett.}, 7\penalty0 (2):\penalty0 5639--5646, April 2022.
\newblock ISSN 2377-3766, 2377-3774.
\newblock \doi{10.1109/LRA.2022.3151613}.

\bibitem[Ngiam et~al.(2021)Ngiam, Vasudevan, Caine, Zhang, Chiang, Ling, Roelofs, Bewley, Liu, Venugopal, Weiss, Sapp, Chen, and Shlens]{ngiam_scene_2021}
Jiquan Ngiam, Vijay Vasudevan, Benjamin Caine, Zhengdong Zhang, Hao-Tien~Lewis Chiang, Jeffrey Ling, Rebecca Roelofs, Alex Bewley, Chenxi Liu, Ashish Venugopal, David~J. Weiss, Benjamin Sapp, Zhifeng Chen, and Jonathon Shlens.
\newblock Scene {{Transformer}}: {{A}} unified architecture for predicting future trajectories of multiple agents.
\newblock In \emph{International {{Conference}} on {{Learning Representations}}}, October 2021.

\bibitem[Park et~al.(2023)Park, Ryu, Yang, Cho, Kim, and Yoon]{park_leveraging_2023}
Daehee Park, Hobin Ryu, Yunseo Yang, Jegyeong Cho, Jiwon Kim, and Kuk~Jin Yoon.
\newblock Leveraging {{Future Relationship Reasoning}} for {{Vehicle Trajectory Prediction}}.
\newblock In \emph{International {{Conference}} on {{Learning Representations}}}, 2023.

\bibitem[Plaat et~al.(2020)Plaat, Kosters, and Preuss]{plaat_deep_2020}
Aske Plaat, Walter Kosters, and Mike Preuss.
\newblock Deep {{Model-Based Reinforcement Learning}} for {{High-Dimensional Problems}}, a {{Survey}}, December 2020.

\bibitem[Rhinehart et~al.(2019)Rhinehart, McAllister, and Levine]{rhinehart_deep_2019}
Nicholas Rhinehart, Rowan McAllister, and Sergey Levine.
\newblock Deep {{Imitative Models}} for {{Flexible Inference}}, {{Planning}}, and {{Control}}.
\newblock In \emph{International {{Conference}} on {{Learning Representations}}}, September 2019.

\bibitem[Ridel et~al.(2020)Ridel, Deo, Wolf, and Trivedi]{ridel_scene_2020}
Daniela Ridel, Nachiket Deo, Denis Wolf, and Mohan Trivedi.
\newblock Scene {{Compliant Trajectory Forecast With Agent-Centric Spatio-Temporal Grids}}.
\newblock \emph{IEEE Robotics and Automation Letters}, 5\penalty0 (2):\penalty0 2816--2823, April 2020.
\newblock ISSN 2377-3766.
\newblock \doi{10.1109/LRA.2020.2974393}.

\bibitem[Ruthotto and Haber(2021)]{ruthotto_introduction_2021}
Lars Ruthotto and Eldad Haber.
\newblock An introduction to deep generative modeling.
\newblock \emph{GAMM-Mitteilungen}, 44\penalty0 (2):\penalty0 e202100008, 2021.
\newblock ISSN 1522-2608.
\newblock \doi{10.1002/gamm.202100008}.

\bibitem[Salzmann et~al.(2020)Salzmann, Ivanovic, Chakravarty, and Pavone]{salzmann_trajectron_2020}
Tim Salzmann, Boris Ivanovic, Punarjay Chakravarty, and Marco Pavone.
\newblock Trajectron++: {{Dynamically-Feasible Trajectory Forecasting}} with {{Heterogeneous Data}}.
\newblock In Andrea Vedaldi, Horst Bischof, Thomas Brox, and Jan-Michael Frahm, editors, \emph{Computer {{Vision}} -- {{ECCV}} 2020}, pages 683--700, Cham, 2020. Springer International Publishing.
\newblock ISBN 978-3-030-58523-5.
\newblock \doi{10.1007/978-3-030-58523-5_40}.

\bibitem[Saxena and Cao(2021)]{saxena_generative_2021}
Divya Saxena and Jiannong Cao.
\newblock Generative {{Adversarial Networks}} ({{GANs}}): {{Challenges}}, {{Solutions}}, and {{Future Directions}}.
\newblock \emph{ACM Comput. Surv.}, 54\penalty0 (3):\penalty0 63:1--63:42, May 2021.
\newblock ISSN 0360-0300.
\newblock \doi{10.1145/3446374}.

\bibitem[Shi et~al.(2022)Shi, Jiang, Dai, and Schiele]{shi_motion_2022}
Shaoshuai Shi, Li~Jiang, Dengxin Dai, and Bernt Schiele.
\newblock Motion {{Transformer}} with {{Global Intention Localization}} and {{Local Movement Refinement}}.
\newblock \emph{Advances in Neural Information Processing Systems}, 35:\penalty0 6531--6543, December 2022.

\bibitem[Sun et~al.(2020)Sun, Kretzschmar, Dotiwalla, Chouard, Patnaik, Tsui, Guo, Zhou, Chai, Caine, Vasudevan, Han, Ngiam, Zhao, Timofeev, Ettinger, Krivokon, Gao, Joshi, Zhang, Shlens, Chen, and Anguelov]{sun_scalability_2020}
Pei Sun, Henrik Kretzschmar, Xerxes Dotiwalla, Aurelien Chouard, Vijaysai Patnaik, Paul Tsui, James Guo, Yin Zhou, Yuning Chai, Benjamin Caine, Vijay Vasudevan, Wei Han, Jiquan Ngiam, Hang Zhao, Aleksei Timofeev, Scott Ettinger, Maxim Krivokon, Amy Gao, Aditya Joshi, Yu~Zhang, Jonathon Shlens, Zhifeng Chen, and Dragomir Anguelov.
\newblock Scalability in {{Perception}} for {{Autonomous Driving}}: {{Waymo Open Dataset}}.
\newblock In \emph{2020 {{IEEE}}/{{CVF Conference}} on {{Computer Vision}} and {{Pattern Recognition}} ({{CVPR}})}, pages 2443--2451, Seattle, WA, USA, June 2020. IEEE.
\newblock ISBN 978-1-72817-168-5.
\newblock \doi{10.1109/CVPR42600.2020.00252}.

\bibitem[Wu et~al.(2020)Wu, Chen, and Metaxas]{wu_motionnet_2020}
Pengxiang Wu, Siheng Chen, and Dimitris~N. Metaxas.
\newblock {{MotionNet}}: {{Joint Perception}} and {{Motion Prediction}} for {{Autonomous Driving Based}} on {{Bird}}'s {{Eye View Maps}}.
\newblock In \emph{2020 {{IEEE}}/{{CVF Conference}} on {{Computer Vision}} and {{Pattern Recognition}} ({{CVPR}})}, pages 11382--11392, Seattle, WA, USA, June 2020. IEEE.
\newblock ISBN 978-1-72817-168-5.
\newblock \doi{10.1109/CVPR42600.2020.01140}.

\bibitem[Xing et~al.(2025)Xing, Feng, Chen, Dai, Hu, Xu, Wu, and Jiang]{xing_survey_2025}
Zhen Xing, Qijun Feng, Haoran Chen, Qi~Dai, Han Hu, Hang Xu, Zuxuan Wu, and Yu-Gang Jiang.
\newblock A {{Survey}} on {{Video Diffusion Models}}.
\newblock \emph{ACM Comput. Surv.}, 57\penalty0 (2):\penalty0 1--42, February 2025.
\newblock ISSN 0360-0300, 1557-7341.
\newblock \doi{10.1145/3696415}.

\bibitem[Xiong et~al.(2025)Xiong, Chen, and Qi]{xiong_fine-grained_2025}
Wenyi Xiong, Jian Chen, and Ziheng Qi.
\newblock Fine-{{Grained Behavior}} and {{Lane Constraints Guided Trajectory Prediction Method}}, April 2025.

\bibitem[Yu and Wang(2021)]{yu_researches_2021}
Liangyao Yu and Ruyue Wang.
\newblock Researches on {{Adaptive Cruise Control}} system: {{A}} state of the art review.
\newblock \emph{Proceedings of the Institution of Mechanical Engineers, Part D: Journal of Automobile Engineering}, 236:\penalty0 095440702110192, May 2021.
\newblock \doi{10.1177/09544070211019254}.

\bibitem[Zare et~al.(2023)Zare, Kebria, Khosravi, and Nahavandi]{zare_survey_2023}
Maryam Zare, Parham~M. Kebria, Abbas Khosravi, and Saeid Nahavandi.
\newblock A {{Survey}} of {{Imitation Learning}}: {{Algorithms}}, {{Recent Developments}}, and {{Challenges}}.
\newblock \emph{IEEE Transactions on Cybernetics}, 54\penalty0 (12):\penalty0 7173--7186, 2023.

\bibitem[Zhao et~al.(2022)Zhao, Zhu, Du, Liao, and Chan]{zhao_novel_2022}
Cong Zhao, Yifan Zhu, Yuchuan Du, Feixiong Liao, and Ching-Yao Chan.
\newblock A {{Novel Direct Trajectory Planning Approach Based}} on {{Generative Adversarial Networks}} and {{Rapidly-Exploring Random Tree}}.
\newblock \emph{IEEE Trans. Intell. Transport. Syst.}, 23\penalty0 (10):\penalty0 17910--17921, October 2022.
\newblock ISSN 1524-9050, 1558-0016.
\newblock \doi{10.1109/TITS.2022.3164391}.

\bibitem[Zhou et~al.(2022)Zhou, Ye, Wang, Wu, and Lu]{zhou_hivt_2022}
Zikang Zhou, Luyao Ye, Jianping Wang, Kui Wu, and Kejie Lu.
\newblock {{HiVT}}: {{Hierarchical Vector Transformer}} for {{Multi-Agent Motion Prediction}}.
\newblock In \emph{Proceedings of the {{IEEE}}/{{CVF Conference}} on {{Computer Vision}} and {{Pattern Recognition}}}, pages 8823--8833, 2022.

\end{thebibliography}

\end{document}